\documentclass[10pt,twocolumn,letterpaper]{article}

\usepackage{iccv}
\usepackage{times}
\usepackage{epsfig}
\usepackage{graphicx}
\usepackage{amsmath}
\usepackage{amssymb}
\usepackage[utf8]{inputenc} 
\usepackage[T1]{fontenc}    
\usepackage{url}            
\usepackage{booktabs}       
\usepackage{amsfonts}       
\usepackage{nicefrac}       
\usepackage{microtype}      
\usepackage{color}         
\usepackage{multirow}
\usepackage[accsupp]{axessibility}

\usepackage[table,xcdraw]{xcolor}
\definecolor{noteback}{RGB}{254,232,200} 
\definecolor{framecolor}{RGB}{252,141,89}
\definecolor{notecolor}{RGB}{215,48,31} 
\usepackage{tcolorbox} 
\usepackage{tikz} 

\usepackage{amssymb,amsthm}
\usepackage{graphics}
\usepackage{bbm,bm}
\usepackage{microtype}
\usepackage[linesnumbered,lined]{algorithm2e}

\usepackage[pagebackref=true,breaklinks=true,letterpaper=true,colorlinks,bookmarks=false]{hyperref}

\iccvfinalcopy 

\def\iccvPaperID{12134} 

\ificcvfinal\fi

\title{Continual Zero-Shot Learning through Semantically Guided Generative Random Walks}
\newcommand{\ones}{\mathbbm 1}
\newcommand{\reals}{\mathbb{R}}

\newcommand{\Expect}{\mathbb{E}}

\newcommand{\Prob}{\mathbb{ P}}

\newcommand{\gzslms}{\text{mSA}}
\newcommand{\gzslmu}{\text{mUA}}
\newcommand{\gzslmh}{\text{mHA}}
\newcommand{\bwt}{\text{BWT}}

\newtheorem{defn}{Definition}[section]

\newtheorem{thm}[defn]{Theorem}

\newtheorem{prop}[defn]{Proposition}
\newtheorem{lem}[defn]{Lemma}
\newtheorem{stat}[defn]{Statement}
\usepackage{amsmath,amsfonts,bm}

\def\eqref#1{equation~\ref{#1}}

\def\1{\bm{1}}

\def\va{{\bm{a}}}

\def\vc{{\bm{c}}}

\def\vx{{\bm{x}}}

\def\vz{{\bm{z}}}

\def\mA{{\bm{A}}}
\def\mB{{\bm{B}}}
\def\mC{{\bm{C}}}
\def\mD{{\bm{D}}}

\def\mI{{\bm{I}}}

\def\mP{{\bm{P}}}

\def\mX{{\bm{X}}}

\DeclareMathAlphabet{\mathsfit}{\encodingdefault}{\sfdefault}{m}{sl}
\SetMathAlphabet{\mathsfit}{bold}{\encodingdefault}{\sfdefault}{bx}{n}

\def\gD{{\mathcal{D}}}

\def\gI{{\mathcal{I}}}

\def\gU{{\mathcal{U}}}

\def\gZ{{\mathcal{Z}}}

\newcommand{\KL}{D_{\mathrm{KL}}}

\DeclareMathOperator*{\argmin}{arg\,min}

\begin{document}


\author{
Wenxuan Zhang$^1$ \qquad
Paul Janson$^{1,2}$ \thanks{Work done during internship at KAUST} \qquad
Kai Yi$^1$ \qquad
Ivan Skorokhodov$^1$ 
\and
Mohamed Elhoseiny$^1$ \\
KAUST$^1$ \quad
University of Moratuwa$^2$
\\
\texttt{\small \{wenxuan.zhang, remond.janson, kai.yi, ivan.skorokhodov, mohamed.elhoseiny\}@kaust.edu.sa}
} 
\maketitle

\begin{abstract}

Learning novel concepts, remembering previous knowledge, and adapting it to future tasks occur simultaneously throughout a human's lifetime. 
To model such comprehensive abilities, continual zero-shot learning (CZSL) has recently been introduced. However, most existing methods overused unseen semantic information that may not be continually accessible in realistic settings.
In this paper, we address the challenge of continual zero-shot learning where unseen  information is not provided during training, by leveraging generative modeling. The heart of the generative-based methods is to learn quality representations from seen classes to improve the generative understanding of the unseen visual space.
Motivated by this, we introduce generalization-bound tools and provide the first theoretical explanation for the benefits of generative modeling to CZSL tasks. Guided by the theoretical analysis, we then propose our learning algorithm that employs  a novel semantically guided Generative Random Walk (GRW) loss. The GRW loss augments the training by continually encouraging the model to generate realistic and characterized samples to represent the unseen space.  Our algorithm achieves state-of-the-art performance on AWA1, AWA2, CUB, and SUN datasets, surpassing existing CZSL methods by 3-7\%. The code has been made available here  \url{https://github.com/wx-zhang/IGCZSL}
\end{abstract}.


\section{Introduction}
\begin{figure*}[t]
  \centering
   \includegraphics[width=0.76\linewidth]{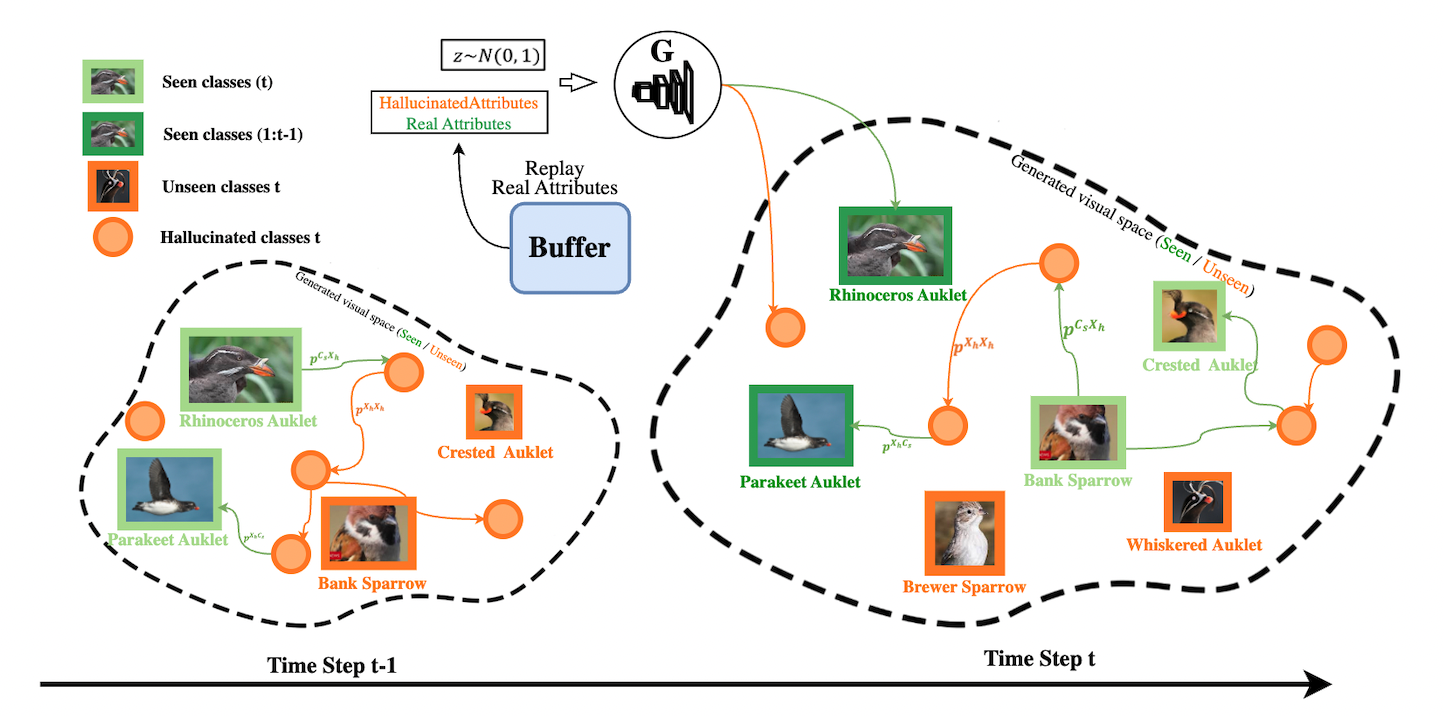}
   \caption{\textbf{Semantically guided generative random walk (GRW)}: At each time step, new classes are added to the seen classes space, and the random walk starts from each seen class center (in \textcolor[rgb]{0,0.7,0}{green}) and transitions through generated samples of hallucinated classes (in \textcolor{orange}{orange}), then the landing probability distribution over the seen classes is predicted. The GRW loss encourages the generated samples from the hallucinated classes to be distinguishable from the seen classes by encouraging the landing probability over seen classes starting 
 from any seen center to be uniformly distributed,  and hence hard to classify to any seen class.}
   \label{fig:introgrw}
   \vspace{-2mm}
\end{figure*}
Researchers have devoted significant effort to developing AI learners to mimic human cognition. One such endeavor is zero-shot learning (ZSL), which aims to identify unseen classes without accessing any of their images during training. However, human zero-shot learning abilities improve dynamically over time. As individuals acquire more knowledge of seen tasks, they become better at recognizing unseen tasks. To evaluate the zero-shot learning in such a dynamic seen-unseen distribution,  the continual zero-shot learning problem (CZSL) has been proposed \cite{skorokhodov2021class}. CZSL emulates the continuous learning process of a human's life, where the model continually sees more classes from the unseen world and is evaluated on both seen and unseen classes.  This CZSL  skill, may it get maturely developed to the world scale, has the potential to accelerate research in species discovery, for example, as known species grow continually, but close to 90\% of the species are not yet discovered\cite{sweetlove2011number}.  

Generative models (e.g., GANs\cite{goodfellow_generative_2014}) have made significant progress in producing photorealistic images by learning high-dimensional probability distributions. This ability  motivated researchers to adapt GANs to ZSL to generate missing data of unseen classes conditioning  on unseen semantic information, known as generative-based ZSL. Training the classifier on synthetic unseen samples can reduce model prediction bias towards seen classes and thus achieves competitive zero-shot learning performance~\cite{li_leveraging_2019,vyas2020leveraging,narayan2020latent}. Some CZSL works directly adopt this framework continually, known as transductive continual zero-shot learning \cite{ghosh_dynamic_2021,kuchibhotla2022unseen}.
However, in CZSL, the unseen world changes dynamically and unexpectedly, making it unrealistic to use prior knowledge about unseen classes\cite{skorokhodov2021class}. When we do not assume access to unseen semantic information in the CZSL setting, which is known as \emph{inductive continual zero-shot learning}, most existing methods struggle to perform well, as we show in our experiments. 
Furthermore, the theoretical understanding of how zero-shot learning benefits from synthetic data is limited, which poses an obstacle to developing purely inductive continual zero-shot methods. 
Recent analyses of training generative models with synthetic data \cite{chang_g2r_2019} provide a possible avenue for developing the desired theoretical explanation. This led us to develop a generalization-bound tool to understand the learning mechanism in generative-based CZSL and further develop inductive methods based on it.

In our analysis, we have identified it is crucial to reduce the distance between the generated and actual visual space of unseen classes. This requires the model to generate realistic samples to represent unseen space to augment the training of the classifier. 
However, the lack of ground truth semantic descriptions for unseen classes and the lack of previously seen classes data often leads to the generated samples collapsing to the seen classes.
A similar problem has been addressed in generating novel style artworks, where GAN training is augmented to encourage the generated styles to deviate from existing art style classes~\cite{can_2017, sbai2018design, hertzmann2018can, djwaga, hertzmann2020visual,jha2022creative}. Drawing inspiration from the improved feature representation achieved by generative models in producing novel art, and the connection between the ability to generate novel styles in art generation and to generate samples to represent the unseen space in generative-based CZSL, we propose a purely inductive, \textbf{Generative Random Walk (GRW)} loss,  guided only by semantic descriptions of seen classes. 

In each continual learning task, we first hallucinate some classes by interpolating on or sampling from a learnable dictionary based on the current and previous classes, with the  belief that the realistic classes, both seen and unseen, should be relatable to each other ~\cite{elhoseiny2019creativity,elhoseiny2021cizslpp}. We then generate samples from the hallucinated classes. To prevent the generated samples of hallucinated classes from collapsing to the seen classes, we apply the GRW loss, as illustrated in Figure \ref{fig:introgrw}. We perform a random walk starting from the \textcolor[rgb]{0,0.7,0}{seen class} and moving through generated examples of  hallucinated classes for $R$ steps, as described in detail later in Section \ref{sec:rw}. The GRW loss encourages high transition probabilities to the \textcolor{orange}{realistic unseen space} by deviating from the visual space of the seen classes and avoiding less realistic areas. The resulting representations are both realistic and distinguishable from seen classes, which enhances the generative understanding of unseen classes. 
This approach is particularly effective when the model is updated continually, as it enables the model to use the newly learned knowledge to improve further the generated examples of  hallucinated classes. 
Our contributions lie in
\begin{itemize}
    \item We provide a theoretical analysis  of continual zero-shot learning. This analysis guides us to use proper signals to make up for the missing unseen information. We present these generalization-bound tools for the analysis in Section \ref{sec:analysis}.
    \item Guided by the analysis, we develop a method for purely inductive continual zero-shot learning; described in detail in Section \ref{sec:method}. Our method, ICGZSL, first provides two ways to hallucinate classes, \ie interpolation of two seen classes and learning a dictionary based on the seen classes.  Then, we integrate our introduced semantically guided Generative Random Walk (GRW) loss to generate distinguishable and realistic samples to represent unseen classes.
    \item We performed comprehensive experiments (Section \ref{sec:exp}) that demonstrate the effectiveness of our approach. Specifically, our model achieves state-of-the-art results on standard continual zero-shot learning benchmarks, AWA1, AWA2, CUB, SUN, and performs often better than transductive methods   
\end{itemize}


\section{Related Works} \label{sec:related}
\textbf{Inductive and Transductive Zero-Shot Learning.} There are varying degrees of accessibility to unseen information in zero-shot learning.  Transductive methods use both unlabeled samples and attributes of unseen classes during training~\cite{paul2019semantically, rahman2019transductive}. Semantically transductive methods, on the other hand, only use attributes of unseen classes in training~\cite{xian_feature_2018, wang_zero-shot_2021}. In the inductive setting, however, no unseen information is allowed to be used(\eg,~\cite{Elhoseiny_2018_CVPR, elhoseiny2019creativity, Liu_2019_ICCV,Xian_2019_CVPR}). This can result in a bias towards seen classes~\cite{pourpanah2020review}. Generative methods, such as those used by \cite{Elhoseiny_2018_CVPR, Liu_2019_ICCV, elhoseiny2019creativity}, can produce unseen samples using only seen class information during training to solve this issue.
For example, 
\cite{elhoseiny2019creativity} relate zero-shot learning to human creativity to generate images that deviate from seen classes during training. \cite{chandhok2021structured} used unlabeled samples from out-of-distribution data to gather knowledge about unseen data.
\cite{Su_2022_CVPR} utilize two variational autoencoders to generate latent representations for visual and semantic modalities in a shared latent space.
In contrast, our approach focuses on investigating the relationship between the generated samples of hallucinated  classes and the seen classes, which leads to GRW loss.

\textbf{Continual Learning. }
The majority of continual learning works aim to tackle the problem of catastrophic forgetting, where the data representation becomes biased towards the most recent task in sequential learning. {Regularization-based methods} ~\cite{li2017learning,aljundi_memory_2018},  
{structure-based methods} ~\cite{rajasegaran2019random,ebrahimi2020adversarial}, and
replay-based methods~\cite{ shin2017continual, xiang2019incremental} have been proposed to resolve this problem. 
More recently, research has explored forward transfer in continual learning, with the belief that as knowledge accumulates, higher next-task transferability, as measured by zero-shot assessment, should be attained.  Their evaluation space either includes the next task~\cite{lin2021clear} or the whole class space~\cite{douillard2021insights}. However, compared to our setting, \cite{lin2021clear} did not evaluate the model in a generalized manner, and \cite{douillard2021insights} only paid attention to the seen accuracy.

\textbf{Continual Zero-shot learning. }
\cite{chaudhry_efficient_2019} introduced A-GEM for continual learning, which was later applied to deal with zero-shot tasks sequentially, laying the foundation for the initial work on CZSL.
 \cite{skorokhodov2021class} proposed the inductive CZSL scenario and demonstrated that a class-based normalization approach can improve performance in continual zero-shot learning.
Both \cite{gautam_generative_2021} and \cite{ghosh_adversarial_2021} explore the CZSL problem, but rely on unseen class descriptions to train a classifier before inference. \cite{kuchibhotla2022unseen} proposed a generative adversarial approach with a cosine similarity-based classifier that supports the dynamic addition of classes without requiring unseen samples for training. Their approach also relies on unseen class descriptions for seen-unseen deviation, making it a semantically transductive method. This motivated us to explore a purely inductive method for handling seen-unseen deviation and improving the realism of unseen samples.

\section{Problem Setup and Notations}
\subsection{Formulation}\label{sec:problemsetup}

We start by defining our problem and notations. 
A labelled dataset is defined as a tuple $ \mD = \{(\vx,\va,y) | y = f(\vx), (\vx, \va, y)\sim \mathcal{D}\}$, where $\mathcal{D}$ represents the data distribution. Each data point is a tuple of image feature $\vx \in \reals^{d_x}$, class attribute $\va  \in \reals^{d_a}$, and a class label $y$.  Here $d_x$ is the dimension of the visual feature space, and $d_a$ is the dimension of the attribute space. Each distribution has a specific labeling function $f$. 
Our goal is to learn a model $\hat{f}$ on top of $\mD$ to estimate $f$. 
We study the continual zero-shot learning setting proposed by \cite{skorokhodov2021class}, where we seek to learn the model $\hat{f}$ on a stream of tasks. In each task $t$, the  model is learned on the seen dataset $\mD^t_s$, and is evaluated on both the seen distribution $\mathcal{D}^t_{s}$ and unseen distribution $\mathcal{D}^t_{u}$. Moreover, we assume that the set of seen class  and unseen class  are disjoint, that is $\mathcal{D}_s\cap \mathcal{D}_u = \phi$. This procedure is illustrated in the bottom part of Figure \ref{fig:approach}. 

 We use generative models as the backbone. During the training time, the model $\hat{f}$ is trained on the seen dataset ${\mD}_s$ as well as the synthesized dataset ${\mD}_{h}$. ${\mD}_{h}$ is generated by conditioning on hallucinated attributes ${\va}_{h}$ and prior $\mathcal{Z} \sim \mathcal{N}(0,1)$. The labeling function ${f}_{h}$ of the generated dataset is a look-up table of the generated features ${\vx} \in {\mX}_{h}$ and the corresponding attribute condition ${\va}_{h}$. 
 
\subsection{Notations.} 
In our theoretical analysis, we use the following notations: 1) We discuss the relationship between the three types of variables, namely, real seen sample,  real unseen samples,  and generated samples from the hallucinated classes. To specify the variables related to these types of samples, we use subscripts $\cdot_s, \cdot_u, \cdot_{h}$ respectively, \eg, $f_s,f_u,f_{h}$  ; 
2) We denote the values and model empirically computed by a variable with a hat, \eg, $\hat{f}$; 3) We use superscripts $\cdot^t$ or $\cdot^{1:t}$ to indicate that a variable is for task $t$ or for tasks $1:t$ respectively, \eg, $f^t_s, f_s^{1:t}$; 4) $\mD$ is used for the empirical sample set, and $\mathcal{D}$ is used for the distribution; 5) We use $N_s$ and $N_u$ to denote the number of seen and unseen classes.

In practice, the unseen information,
\ie, $\va_u, \mD_u, N_u$, is not available. Therefore, we hallucinate some classes denoted by $\va_h$ and generate samples $\mD_h = \{(\vx_h, \va_h)\}$ by conditioning on these attributes. We use $N_h$ to represent the number of hallucinated classes. 
Additionally, we do not have access to all the previous data
, so $\mX^{1:t}$ refers to the current samples as well as the previous ones in the buffer. 
We also use generated seen samples $\cdot_{sg}$ for GAN training.

\section{Theoretical Analysis} \label{sec:analysis}
As mentioned in the introduction, we propose using hallucinated classes to represent the unseen space. By training our model on synthetic samples generated from these classes, we improve the model's generalization ability to the actual unseen classes during the testing time of continual zero-shot learning. In this section, we quantify the model's generalization ability by measuring the distance between the synthetic samples that represent the unseen space and the actual unseen samples. Additionally, we explain our motivation for using a random walk-based method to reduce this distance when no information about the unseen space is available. 

\subsection{Generalization Bound Inductive Continual Zero-Shot Learning} \label{sec:bound}
In this section, we present a generalization bound for a continual zero-shot learning algorithm. Given the entire training distribution, a learning algorithm can output an optimal hypothesis $h$ that estimates the ground truth labeling function $f$. However, since the learning algorithm can only be trained on a finite sample from the training set, it outputs an empirical hypothesis $\hat{h}$ to estimate the ground truth labeling function. We define the generalization error~\cite{kearns1994introduction} for these two types of hypotheses.
We define the actual risk, 
\begin{equation}\small
    \epsilon(h,f) = \Expect_{(\vx,\va) \sim \mathcal{D}}[\ones_{f(\vx) \neq h(\vx,\va)}] \enspace,
\end{equation}
which measures the expected probability of a disagreement between the ground truth and the optimal hypothesis. We also define the empirical risk on the finite sample set $\mD$, 
\begin{equation}\small
    \hat{\epsilon}(\hat{h},f) = \frac{1}{|\mD|}\sum_{(\vx,\va) \in \mD}\ones_{f(\vx) \neq \hat{h}(\vx,\va)} \enspace,
\end{equation}
which measures the probability of a disagreement between the ground truth and the empirical hypothesis.

In a continual zero-shot learning algorithm, given a training set $\mD^t_s$, the algorithm outputs $\hat{h}$ to estimate $f^{1:t}_s \cup f_u$ instead of the ground truth labeling function $f_s$. To begin our analysis, we propose a distance measure between the generated unseen distribution\footnote{In the transductive setting, the unseen distribution is generated by conditioning on the unseen semantic information. In our work, we utilize generated samples from hallucinated classes to represent the generated unseen distribution.} and the real unseen distribution $\bar{d}_{GDB}(\mD_{h}, \mD_u)$ as follows:
\begin{defn}[Empirical Generative distance]
\label{defn:1}
Given the training set  $\mD_s$ and the synthetic set $\mD_{h}$  
, the ground truth labeling functions $f_s$, $f_{h}$, and $f_u$, and the optimal hypothesis $\hat{h}^* = \argmin_{h\in H}\hat{\epsilon}_s(h,f_s) + \hat{\epsilon}_{h}(h,f_{h})$ obtained by training the model on $\mD_{h}$ and $\mD_s$, we can define the distance between $\mD_{h}$ and $\mD_u$ as follows:
\begin{equation} \label{eq:defgdb}\small
    \bar{d}_{GDB}(\mD_{h}, \mD_u) = |\hat{\epsilon}(\hat{h}^*,f_u) - \hat{\epsilon}(\hat{h}^*,f_{h})| \enspace.
\end{equation}
\end{defn}
Our proposed $\bar{d}_{GDB}$ is a feasible distance measure that satisfies the properties of a pseudo-metric. In the following, we present our generalization bound following \cite{chang_g2r_2019} for the continual zero-shot learning algorithm, which shows how the generalization ability of the zero-shot learning algorithm is mainly influenced by this distance.
\begin{thm}[Generalization bound of the generative-based CZSL]\label{thm:gbzsl}
Given the CZSL procedure described in section \ref{sec:problemsetup}, with confidence $1-\delta$ the risk on the unseen distribution is bounded by 
\begin{equation} \label{eq:main}\footnotesize
\begin{split}
    \epsilon(h, f_u^t) \leq  &\hat{\epsilon}(\hat{h}^*,f_{s}^{1:t})  + \frac{1}{2}d_{\mathcal{H}\Delta \mathcal{H}}(\mathcal{D}_{s}^{1:t},\mathcal{D}^t_u)+ \bar{\lambda} \\
    &+ \frac{1}{2} \bar{d}_{GDB}(\mD^t_u, \mD^t_{h}) + C (\frac{1}{m} + \frac{1}{\delta})
\end{split}
\end{equation}
where $\hat{h}^* = \argmin_{h\in H}\sum_{i=1}^t\hat{\epsilon}(h,f_{s}^i) + \hat{\epsilon}(h,f^t_{h})$, $\bar{\lambda} = \hat{\epsilon}(\hat{h}^*,f_{s}^{1:t}) + \hat{\epsilon}(\hat{h}^*,f^t_{h})$. 
\end{thm}
In Equation~\ref{eq:main}, measurement $d_{\mathcal{H}\Delta \mathcal{H}}$~\cite{ben-david_theory_2010} is used to quantify the difference between two distributions for domain adaptation based on the type of model, and
 is fixed for a specific problem.  $\bar{\lambda}$ and $\hat{\epsilon}_s(h,f_s)$ are highly depended on the optimization algorithm. However, if we hallucinate a diverse set of classes, $f_u$ can be compactly supported by $f_{h}$. If we further generate realistic samples for each of the hallucinated classes, the optimal solution trained on the synthetic set, $\hat{h}^* = \argmin_{h\in H}\hat{\epsilon}_s(h,f_s) + \hat{\epsilon}_{h}(h,f_{h})$, should perform well on  the real unseen dataset.   This can lead to a reduction in $\bar{d}_{GDB}(\mD_u, \mD_{h})$ in Equation~\ref{eq:defgdb}. We will discuss this further in the following section.
The detailed derivation of this theorem can be found in Appendix ~\ref{appen:thm1}.

\subsection{Reducing the bound using Markov Chain.}
\label{subsec:reducing-bound}\label{sec:connection}
To reduce $\bar{d}_{GDB}(\mD_{h}, \mD_u)$ in Equation \ref{eq:main}, we need to decrease the difference between $\mD_u$ and $\mD_{h}$. One approach proposed by \cite{elhoseiny2019creativity} is to hallucinate $\va_{h}$ as a compact support of $\va_u$. Once we have achieved this, we can further generate high-quality samples to increase $\Prob[\mD_u \subset \mD_{h}]$, where the probability is taken over all possible generations.

To quantify the probability value $\Prob[\mD_u \subset \mD_{h}]$, we follow the approach of \cite{van2017unsupervised} and view the generations as nodes in a Markov chain. We define the transition probability between two states as the probability with which one sample is classified as another. Then, we can bound $\Prob[\mD_u \subset \mD_{h}]$ by the self-transition probability using a generalization bound.
When the self-transition probability is the same in two sets of generations, we prefer the one with higher diversity quantified by DDP, as suggested by \cite{kang2013fast} and \cite{elfeki2019gdpp}.

For detailed explanations, please refer to Appendix ~\ref{appen:stat}. Here, we provide an informal statement.
\begin{stat} \label{stat:main}
    Finding generated samples from hallucinated classes to ``carefully'' increase the determinant and the diagonal entries of the transition matrix of the above described Markov Chain can reduce $\bar{d}_{GDB}$.
\end{stat}

We can now design an algorithm that first hallucinates classes and then generates diverse samples from these classes to represent the unseen space that follows Statement \ref{stat:main}. However, the transition matrix of the Markov chain described above is intractable to compute in practice. To quantify the transition probability, we adapt the random walk framework~\cite{8099557, ayyad2020semi} originally used in semi-supervised few-shot learning to generative zero-shot learning with a few yet important changes. Please refer to Appendix ~\ref{appensec:relation} the relation between our work and previous work. 

We also make the following two adjustments to the statement
to encourage the generated samples from hallucinated classes to be consistently realistic like the real samples. Firstly, we represent the transition matrix among hallucinated classes (noted as $\mP^{X_hX_h} \in \reals^{N_h\times N_h}$) in the seen class space using a congruent transformation $\mP^{\mC_s X_h} \mP^{X_hX_h}\mP^{ X_h C_s }$, where $\mP^{\mC_s X_h}\in \reals^{N_s \times N_h}$ is the transition probability matrix from seen prototypes to generated samples from hallucinated classes, and $\mP^{ X_h C_s }$ is the opposite.
Secondly, hallucinating a compact support of unseen class attributes and encouraging the transition matrix to be diagonal requires a huge number of generations. To reduce this number, we encourage the generated samples of hallucinated classes to have a "relatable deviation" to the seen classes. The relationship between the two types of samples is that both should be realistic. This means that the transition matrix $\mP^{X_hX_h}$ may not be strictly diagonal, and our goal is to reduce the non-diagonal entries, \ie, to reduce the transition probability between different generated samples of hallucinated classes. We further repeat the transition among generated samples of hallucinated classes to further reduce the non-diagonal entries.

In conclusion, our transitions  start from the seen prototypes to generated samples of hallucinated classes for ${R}$ steps and back to seen prototypes, the transition matrix of which is $\mP^{\mC_s X_h} ({\mP^{X_hX_h}})^R\mP^{ X_h C_s } \in \reals^{N_s \times N_s}$. To encourage``relatable deviation'' of the  generated samples of hallucinated classes  from seen classes, we aim to reduce the non-diagonal entries of the transition matrix 
, as detailed later in Section \ref{sec:method}. This approach intuitively prevents the generations from being attracted by any seen classes, and theoretically can reduce the distance $\bar{d}_{GDB}$. Intentionally, this method also transfers knowledge between seen and hallucinated classes, which is useful for generating realistic images.


\section{Generative-based Inductive CZSL Approach} \label{sec:method}
\textbf{Method overview.} Generative-based inductive CZSL algorithms adopt generative models as their architecture, where seen samples are used to train the classifier to correctly classify seen classes, and generators are trained to generate realistic samples. At the same time, the generator is encouraged to synthesize samples to represent unseen classes to train the classifier to perform classification on these samples. In our work, we can only hallucinate some classes to represent the actual unseen space and generate samples from the hallucinated classes. As guided by our analysis, the key point of inductive zero-shot learning is to generate realistic and diverse samples from hallucinated classes that are deviated from the real seen space. We introduce the generative model backbone in Section \ref{sec:backbone}, and how we generate the abovementioned samples in Section \ref{sec:ag}. The overall procedure is shown in Algorithm  \ref{alg:main} in Appendix.

\subsection{Generative-based CZSL baseline}\label{sec:backbone}
\begin{figure}
    \centering
    \includegraphics[width=\linewidth]{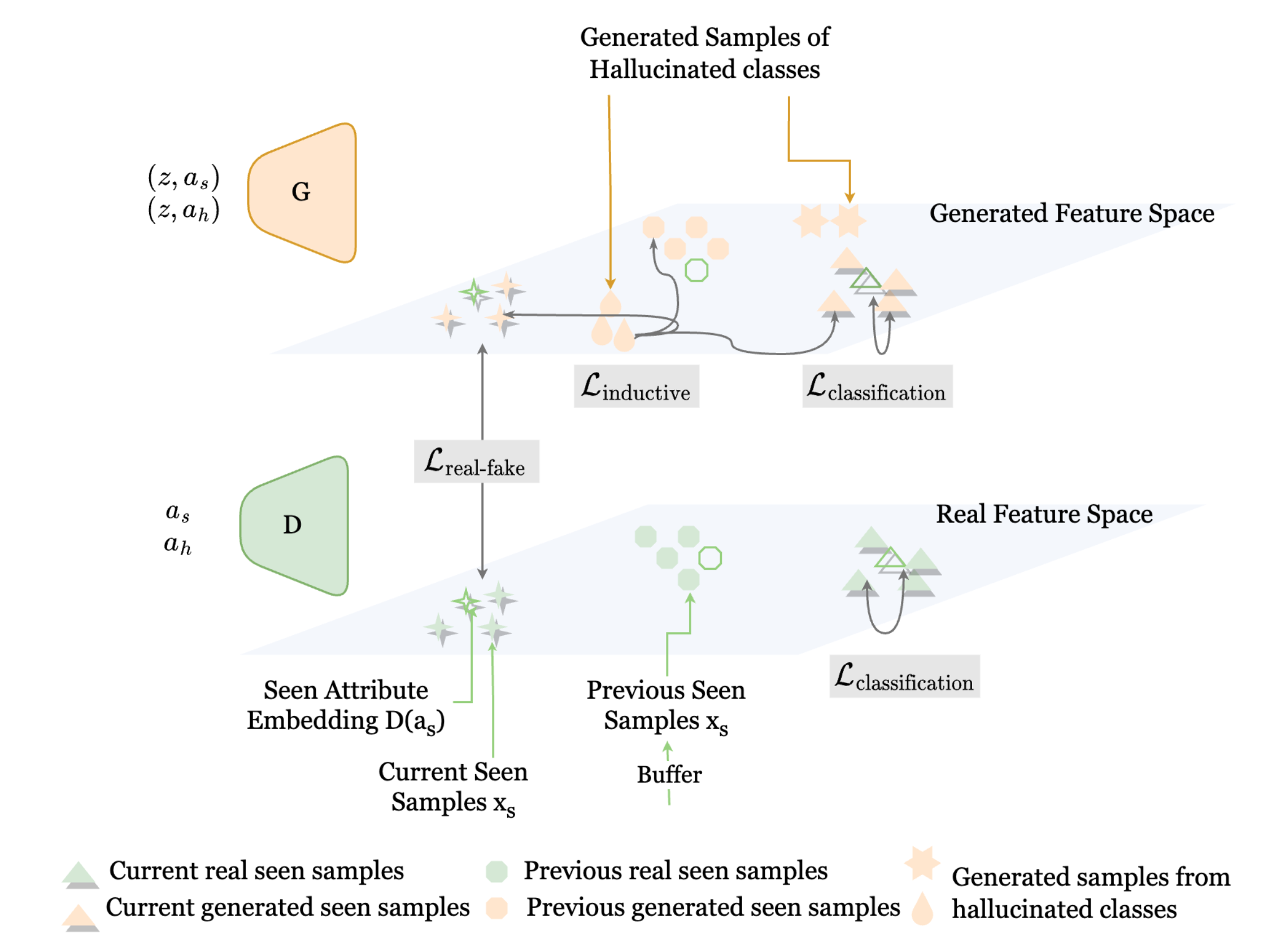} 
    \caption{The discriminator embeds attributes $\va_s^{1:t}$ into the real feature space to perform classification with real samples $x_s^{t}$, while the generator produces features $\vx^t_{h}$ and $\vx^t_{sg}$ conditioning on the corresponding attributes.  The real-fake loss and classification loss encourage  the generated sample distribution consistent with the real samples.  Then the inductive loss applied to the generated feature space, which encourages the characterization of generated samples from hallucinated classes, can reduce the bias towards the current seen classes  of the classifier and improve continual zero-shot learning performacne. }
    \label{fig:approach}
\end{figure}
We follow \cite{kuchibhotla2022unseen} as our baseline. The model contains a generator $G(\va,\vz): \mathbb{R}^{d_a + d_z} \to \reals^{d_x}$ and a discriminator $D(\va)  : \reals^{d_a} \to \mathbb{R}^{d_x}$.
The generator takes the semantic information (denoted by $\va$) and the prior (denoted by $\vz$) sampled from a standard normal distribution $\gZ$ as input and outputs visual features.  Discriminator projects semantic information $\va$ into visual space. 
The conditional adversarial training can be illustrated by the discriminator loss and generator loss as:
\begin{equation}\small
\label{eq:discriminator}
    \begin{split}
     \small
    \mathcal{L}_D &= -\mathcal{L}_\text{real-fake} + \lambda_\text{cls}\mathcal{L}_\text{classification} 
    + \lambda_\text{rd}\mathcal{R}_D , \, \, \,\, \\ 
     \mathcal{L}_G &=  \mathcal{L}_\text{real-fake} + \lambda_\text{cls}\mathcal{L}_\text{classification} + \mathcal{L}_\text{inductive} + \lambda_\text{rg}\mathcal{R}_G . \\
    \end{split}
\end{equation}
As shown in Figure \ref{fig:approach}, we use $\mathcal{L}_\text{real-fake}$ to denote the GAN loss that discriminates between the real and fake samples for the current task, and $\mathcal{L}_\text{classification}$ to denote the entropy loss based on cosine similarity that is used to perform the classification of all seen classes up to the current task. The equations for $\mathcal{L}_\text{real-fake}$ and $\mathcal{L}_\text{classification}$ are shown below
\begin{equation}\label{eq_adv}\footnotesize
\begin{split}
    \mathcal{L}_\text{real-fake}  & = \Expect_{(\vx,\va) \sim \mD^t_s}\big[\log\langle \vx, D(\va)\rangle\big] \\
&-\Expect_{\vz \sim \mathcal{Z}, (\va,\vx) \sim \mD^t_s} \big[\log\langle G(z,\va), D(\va)\rangle\big] \\
     \mathcal{L}_\text{classification} &=
      \Expect_{(\vx, y) \sim \mD^{1:t}_s} \big[L_{e}(\langle \vx, D(\mA_s^{1:t})\rangle,y)\big] \\ 
      &+ \Expect_{{\vz \sim\mathcal{Z},(\vx,\va, y) \sim \mD^{1:t}_s} } \big[L_{e}(\langle G(\vz,\va), D(\mA_s^{1:t})\rangle,y) \big]
    \end{split}
\end{equation}
where $\langle\cdot,\cdot\rangle$ represents the cosine similarity, $\mA_s^{1:t}$ is a matrix of attributes of seen classes up to the current task, and $L_e$ is the cross-entropy loss. In practice, $\mD^{1:t}_s$ consists of the current samples and previous samples in the buffer. We follow \cite{kuchibhotla2022unseen} for regularization terms $ \mathcal{R}_D, \mathcal{R}_G$ and $\lambda_{c,rd, rg}$. See  Appendix ~\ref{appen:fullalg}  for more details about the baseline algorithm.  $\mathcal{L}_\text{inductive}$ with its corresponding $\lambda_i$ is the main component to improve inductive continual zero-shot learning, which will be described in detail in section \ref{sec:ag}.

\subsection{Inductive Loss}\label{sec:ag}

\subsubsection{Hallucinate Attributes} \label{sec:attr_gene}
To begin our method, we first hallucinate classes to represent the unseen space. During this procedure, we aim to generate a diverse and compact set of attributes without using any information from the unseen test set.

\textbf{Interpolation-based method.}
When the attributes are distributed uniformly in the attribute space, which can be compactly supported by the seen attribute, we use interpolation-based method. To hallucinate the  attributes at every mini-batch, we use an interpolation-based method that was introduced by \cite{elhoseiny2019creativity}. Hallucinated attributes are generated using the formula $\va_{ug} = \alpha \va_{s_1} + (1-\alpha) \va_{s_2}$, where $\alpha$ is drawn from the uniform a distribution $\mathcal{U}(0.2,0.8)$, and $\va_{s_1}$ and $\va_{s_2}$ are two randomly chosen seen attributes. The sample interval is chosen to be $(0.2,0.8)$ to ensure that the interpolated attributes are not too close to the seen attributes.

\textbf{Dictionary-based method.} We further propose to learn an attribute dictionary containing $N_s^t$ attribute vectors during training. The use of a learnable dictionary allows the attributes to change more freely in accordance with the loss function. The dictionary is randomly initialized by interpolating seen attributes, and during the computation of GRW loss, we randomly pick attributes from it. This approach is particularly useful for classification at a finer level, where the attributes are more specific.

If the hallucinated class can accurately represent the actual unseen space, which is only accessible during the test time, then the model will have good generalization ability on the test set. We visualize the hallucinated classes to examine if this assumption holds. Please refer to  Appendix ~\ref{appen:mehtodvisual} for the visualization of our hallucinated attributes.

\begin{table*}[t]\footnotesize
\centering
\begin{tabular}{lccccccccccccccc}
\toprule
Dataset & \multicolumn{3}{c}{AWA1} &  & \multicolumn{3}{c}{AWA2} &  & \multicolumn{3}{c}{CUB} &  & \multicolumn{3}{c}{SUN} \\ \cmidrule{2-4} \cmidrule{6-8} \cmidrule{10-12} \cmidrule{14-16} 
Metric & mSA & mUA & mHA &  & mSA & mUA & mHA &  & mSA & mUA & mHA &  & mSA & mUA & mHA \\ \midrule
EWC (\textit{cl})~\cite{doi:10.1073/pnas.1611835114}  & 29.4 &	9.0 &	13.8 &&	30.8 &	10.5 & 15.8 &&	12.2 &	0.8 &	1.3 &&	11.6 &	2.6 &	4.1 \\
A-GEM (\textit{cl})~\cite{chaudhry_efficient_2019} & 64.2 &	3.9 &	7.2&&	65.8&	6.7&	11.9&&	14.4&	0.4&	0.8&&	8.6&	3.0&	4.2\\ \midrule
Tf-GZSL (\textit{tr})~\cite{gautam2022tf} & 70.8  & 27.4  & 37.9  && 78.6  & 28.7  & 41.1  && 46.3  & 30.8  & 35.3  && 15.3  & 30.7  & 18.7\\
DVGR (\textit{tr})~\cite{ghosh_dynamic_2021} 
& 65.1 & 28.5 & 38.0 &  & 73.5 & 28.8 & 40.6 &  & 44.9 & 14.6 & 21.7 &  & 22.4 & 10.7 & 14.5 \\
A-CGZSL (\textit{tr})~\cite{ghosh_adversarial_2021} 
& 71.0 & 24.3 & 35.8 &  & 70.2 & 25.9 & 37.2 &  & 34.3 & 12.4 & 17.4 &  & 17.2 & 6.3 & 9.7 \\
BD-CGZSL (\textit{tr})~\cite{kuchibhotla2022unseen} 
& 62.9 & 29.9 & 39.0 &  & 68.1 & 33.9 & 42.9 &  & 19.8 & 17.2 & 17.8 &  & 27.5 & 15.9 & 20.0 \\ \midrule
CN-CZSL (\textit{in})~\cite{skorokhodov2021class} 
& - & - & - &  & 33.6 & 6.4 & 10.8 &  & 44.3 & 14.8 & 22.7 &  & 22.2 & 8.2 & 12.5 \\
BD-CGZSL-in (\textit{in})~\cite{kuchibhotla2022unseen} 
  & 62.1 & 31.5 & 40.5 &  & 67.7 & 32.9 & 42.3 &  & 37.8 & 9.1 & 14.4 &  & 34.9 & 14.9 & 20.8 \\ 
CARNet (\textit{in})~\cite{gautam2022refinement} &	67.6&	27.4&	37.0&&-&-&-&&				42.4&	12.4&	18.8&&	31.5&	15.9&	20.9 \\
\midrule
\textbf{ours} + interpolation & 67.0 & 34.2 & \textbf{43.4} &  & 71.1 & 34.9 & 44.5 &  & 42.2 & 22.7 & {28.4} &  & 36.0 & 21.6 & 26.8 \\
\textbf{ours} + dictionary & 67.1 & 33.5 & 41.6 &  & 70.2 & 35.1 & \textbf{44.6} &  & 42.4 & 23.6 & \textbf{28.8} &  & 36.5 & 21.8 & \textbf{27.1} \\ \bottomrule
\end{tabular}%
\caption{Our proposed method achieves state-of-the-art results when compared with traditional continual learning method (\textit{cl}) recent inductive (\textit{in}) methods and even shows competitive results in mHA with recent semantic transductive methods (\textit{tr}).}
\label{tab:main}
\end{table*}
\subsubsection{Improve Generation Quality by Inductive Loss}\label{sec:rw}
As we discussed in Section \ref{subsec:reducing-bound}, we use GRW loss to improve the generation quality such that the generated samples are realistic, diverse and characterized.  To encourage diversity of the samples generated from the hallucinated attributes, we firstly generate only one sample for each hallucinated attribute. And then, we perform a random walk to compute the transition probability using generated seen samples $\mX_{sg}$ and generated samples from hallucinated samples $\mX_{h}$. 
The random walk starts from each generative seen class center $\mC_s \in \reals^{N_s^{1:t}\times d_x}$  computed by the mean of generated seen samples from the corresponding class attributes, where $N_s^{1:t}$ are the number of seen classes until step $t$. Then we take $R$ steps of transitions within generated samples of hallucinated classes $\mX_h$ with the final landing probability over seen classes so far. 
The transition probability matrix from seen class centers to generated samples of hallucinated classes  is defined as
\begin{equation}\small
    \mP^{C_s X_{h}} = \sigma(\langle \mC_s ,\mX_{h}^\top\rangle) \enspace,
\end{equation}
where $\langle \cdot, \cdot \rangle$ is a similarity measure, and $\sigma(\cdot)$ is a softmax operator applied on rows. In practice, we use negative Euclidean distance for similarity, that is, suppose $\vx_{h}$ is the row  $i$ of $\mX_{h}$ and $\vc$ is the  class center $j$, 
\begin{equation}\small
  \langle \mC_s, \mX_{h}^\top\rangle_{i,j} = -\|\vx_{h} - \vc\|^2 \enspace.  
\end{equation}
Similarly, the transition probability matrix within generated samples of hallucinated classes and from generated samples of hallucinated classes to seen class centers are defined as
\begin{equation}\small
    \mP^{X_hX_h} = \sigma(\langle \mX_{h}, \mX_{h}^\top\rangle), \mP^{ X_h C_s } = \sigma(\langle \mX_{h}, \mC_s^\top\rangle)\enspace.
\end{equation}
Then the random walk staring from each seen class center and transiting $R$ steps within generated samples of hallucinated classes and back to seen centers are computed by
\begin{equation} \label{eq:single_rw}\footnotesize
    P^{\mC_s \mX_{h} \mC_s}(R) = \mP^{C_s X_h} ({\mP^{X_hX_h}})^R\mP^{ X_h C_s }  \enspace
\end{equation}
In practice, we set the diagonal values of $\mP^{X_hX_h} $  to small values and hope to reduce the non-diagonal values. This equals to encourage the probability $P^{\mC_s \mX_{h} \mC_s}(R)$ to be  uniformly distributed over all the seen classes. We further encourage the probability $\mP^{C_s X_{h}}\in \reals^{N_s^{1:t} \times N_h}$ to be uniformly distributed over all the generated  examples to encourage as many generations to be visited in the random walk, and hence encourage the diversity. Hence,  our {\it Generative Random Walk }(GRW) loss is defined by 
\begin{equation} \small
{L}_\text{GRW} =  \sum_{r=0}^{R} {\gamma^{r} L_e(P^{\mC_s \mX_{h} \mC_s}(r),\textcolor{orange}{\gU})}
  + L_e( \mP_v(C_s, X_{h}),\gU_v) \enspace,
\label{eq:RW_loss}
\end{equation}
where $L_e(\cdot,\cdot)$ is the cross-entropy loss, $\gU \in  \reals^{N_s^{1:t} \times 
N_s^{1:t}}$ is uniform distribution,   $R$ is the transition steps, and  $\gamma$ is exponential decay.  We compute  the  probability that each generated point be visited by any  seen class as $P_v(C_s,X_h) = \frac{1}{\tilde{N_s^{1:t}}} \sum_{i=0}^{N_s^{1:t}} \mP^{C_s X_h}_i$, where $\mP^{C_s X_h}_i$ represents the $i^{th}$ row of the $\mP^{C_s X_h}$ matrix. The visit loss is then defined  as the  cross-entropy between $P_v$ and the uniform distribution $\gU_v \in \reals^{N_h}$, encouraging all the generated examples to be visited.  
\label{sec:numerical}
In addition, we empirically found that the GRW loss can also work as a regularizer to encourage the consistency of generated seen visual space as well, which we defined as
\begin{equation}\label{eq:rwreg} \footnotesize
    {R}_\text{GRW} = \sum_{r=0}^{R} {\gamma^{r} L_e(P^{\mC_s \mX_{sg} \mC_s}(r),\textcolor[rgb]{0,0.7,0}{\gI})}
  + L_e( \mP_v(C_s, X_{sg}),\gU_v) ,
\end{equation}
where $\gI$ is identity distribution, and $\mD_{sg}$ represents the matrix for generated seen samples. 

We numerically show that the random walk-based penalty can reduce $\Bar{d}_{GDB}$ (Def \ref{defn:1}) by the relationship between $\Bar{d}_{GDB}$ and $ {L}_{GRW} $.  Details are shown in Appendix ~\ref{appen:numericalgrw}.

We also adapt the loss proposed in~\cite{elhoseiny2019creativity} to directly prevent the generated unseen samples from being classified into seen classes, i.e.,
\begin{equation} \label{eq:cretivity} \footnotesize
    L_\text{creativity} =\Expect_{\vz\sim \mathcal{Z}, \va_{h}\sim \mD_{h}}
  \KL \big(\big\langle G(\vz,\va_{h}), D(\mA_s^{1:t})\big\rangle \Vert \gU \big) ,
\end{equation}
where  $\KL(\cdot\Vert\cdot)$ is the KL divergence,  $\mA_s^{1:t} \in \mathbb{R}^{N_s^{1:t} \times d_a}$ is the matrix of seen classes attributes vectors  until task $t$,     $\va_{h}$ is hallucinated attributes according to Section \ref{sec:attr_gene},  $\big \langle G(\vz,\va_{ug}), D(\mA_s^{1:t})\big\rangle \in \mathbb{R}^{ N_s^{1:t}}$ are the logits over seen classes so far for a given $G(\vz,\va_{h})$, $\gU$ is the uniform distribution.

\textbf{Inductive loss}
Combining Equation \ref{eq:RW_loss}, \ref{eq:rwreg}  and \ref{eq:cretivity} our final inductive loss is
\begin{equation}\small
    \mathcal{L}_\text{inductive} = \lambda_\text{c}L_\text{creativity} + \lambda_\text{i}{L}_\text{GRW} + \lambda_\text{i}{R}_\text{GRW}
\end{equation}
where $\lambda_i$ is the scaling weight for both the GRW loss term and regularization term. 


\section{Continual Zero-Shot Learning Experiment} \label{sec:exp}
\subsection{Experiment Setup}

\textbf{Data Stream and Benchmarks:} We adopt the continual zero-shot learning framework proposed in \cite{skorokhodov2021class}. In this setting, a $T$-split dataset $D^{1:T}$ forms $T-1$ tasks. At time step $t$, the split $D^{1:t}$ is defined as a seen set of tasks, and the split $D^{t+1:T}$ is an unseen set of tasks. We conduct experiments on four widely used CGZSL benchmarks for a fair comparison: AWA1~\cite{lampert2009learning}, AWA2~\cite{yu_semantic_2020}, Caltech UCSD Birds 200-2011 (CUB)\cite{WahCUB_200_2011}, and SUN\cite{patterson_sun_2012}. We follow \cite{skorokhodov2021class, kuchibhotla2022unseen} for the class split in the continual zero-shot learning setting. More details can be found in Appendix D.

\textbf{Baselines, backbone, and training:} We use the method proposed in \cite{kuchibhotla2022unseen} as the main baseline and compare it with recent CGZSL methods in the setting we mentioned above, including the transductive method Tf-GZSL~\cite{gautam2022tf}, DVGR~\cite{ghosh_dynamic_2021}, A-CGZSL~\cite{ghosh_adversarial_2021}, BD-CGZSL~\cite{kuchibhotla2022unseen}, and the inductive method CN-CZSL~\cite{skorokhodov2021class}, CARNet~\cite{gautam2022refinement}. `BD-CGZSL-in' denotes our modified inductive version of \cite{kuchibhotla2022unseen} by naively removing unseen information. Following ~\cite{kuchibhotla2022unseen}, we also compare our baseline with the classical continual learning methods EWC~\cite{doi:10.1073/pnas.1611835114} and A-GEM~\cite{chaudhry_efficient_2019}. We use vanilla GAN's Generator and Discriminator, both of which are two-layer linear networks. Image features are extracted by ResNet-101, pre-trained on ImageNet 1k. The are attributes from \cite{gbu} and extracted features are used as our model input. We use a replay buffer with a fixed size of 5k.

We run all experiments for 50 epochs and 64 batch sizes with the Adam optimizer. We use a learning rate of 0.005 and a weight decay of 0.00001. Results reported in Table \ref{tab:main} are based on one NVIDIA Tesla P100 GPU. We select our random walk steps $R$, weight decay $\gamma$ and coefficient of inductive loss terms $\lambda_i$ according to prior exploratory zero-shot learning experiments shown in Appendix C.

\textbf{Metrics:}
We use the mean seen accuracy, mean unseen accuracy and mean harmonic seen/unseen accuracy~\cite{skorokhodov2021class} to measure the zero-shot learning ability. These metrics are defined as follows, 
\begin{equation}\footnotesize
\begin{split}
    &\gzslms = \frac{1}{T}\sum_{t=1}^{T} S_t(D^{1: t}),  \gzslmu = \frac{1}{T-1}\sum_{t=1}^{T-1}U_t(D^{t+1:T}) \\
    &\gzslmh =\frac{1}{T-1}\sum_{t=1}^{T-1} H(S_t(D^{1: t}), U_t(D^{t+1:T})),
\end{split}
\end{equation}
where  $H(\cdot,\cdot)$ is the harmonic mean and $S_t, U_t$ are seen and unseen per-class accuracy using the model trained after time $t$. We also use the backward transfer~\cite{chaudhry_efficient_2019,yi_domain-aware_2021,skorokhodov2021class} to measure the continual learning ability, which is defined in \cite{skorokhodov2021class}
\begin{equation}\footnotesize
\bwt = \frac{1}{T-1}\sum_{t=1}^{T-1} (S_T(D^{1: t}) - S_t(D^{1: t}) ) \enspace.
\end{equation}
Note that this should only be conducted on seen set, since part of the early unseen set become seen set later. The BWT on unseen set cannot reflect the knowledge retain ability of the model. 

\subsection{Results }
\begin{figure}[t]
  \centering
   \includegraphics[width=1.0\linewidth]{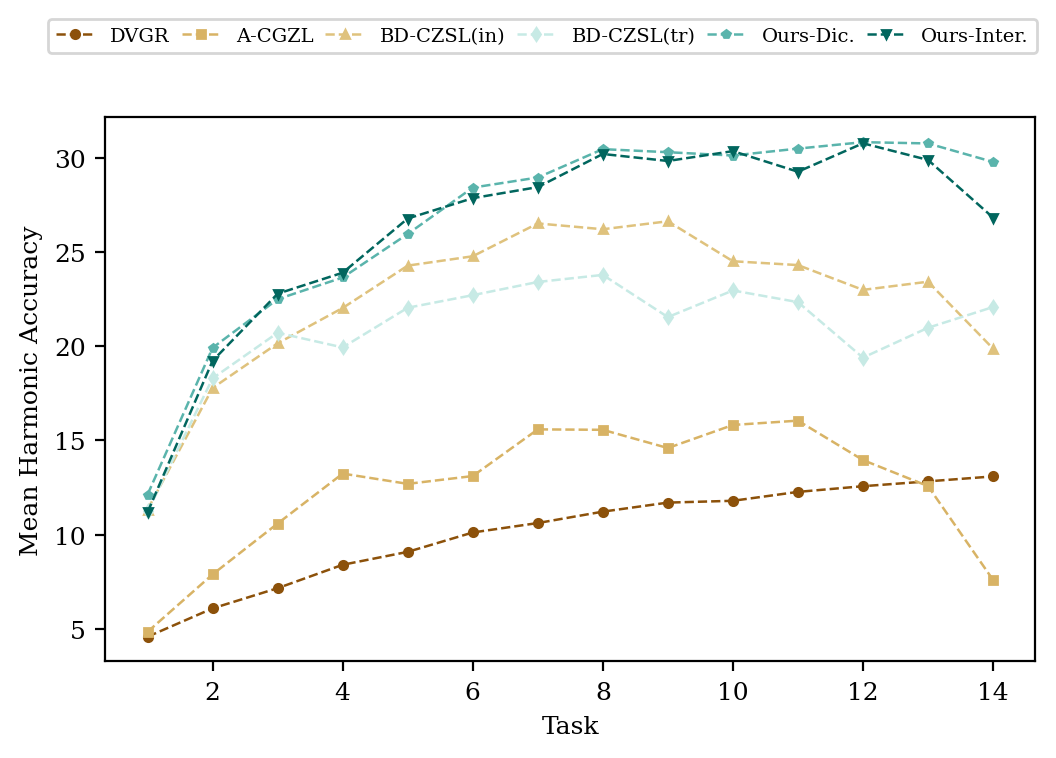}
    \caption{Mean Harmonic accuracy up to each task on SUN dataset. Our  method outperforms both transductive  and inductive methods}
    \label{fig:main}
\end{figure}

The mean harmonic accuracy of the four benchmarks is shown in Table \ref{tab:main}, and the task-wise mHA of the CUB dataset is shown in Figure \ref{fig:main}. In  coarse-grained datasets AWA1 and AWA2, our proposed learner achieves $43.4\% $ and $44.6\%$ in mHA, respectively, surpassing all the current inductive and transductive methods. In the fine-grained datasets and tasks with long steps (CUB, SUN), our method achieves $28.8\%$ and $27.1\%$, surpassing all the current CZSL methods.
 We observe that even though other methods have comparable mSA, they have far lower mUA than ours. 
We believe that our method achieves this improved knowledge transfer ability from seen visual space to unseen visual space through the proposed inductive learning signals, \ie, $\mathcal{L}_\text{inductive}$.
\begin{table}[b]\centering\footnotesize
\begin{tabular}{lccccc}
\toprule
Dataset &  & AWA1 & AWA2 & CUB & SUN \\ \midrule
DVGR~\cite{ghosh_dynamic_2021} & tr & 0.09 & 0.10  & -0.07  & -0.20 \\
A-CGZSL~\cite{ghosh_adversarial_2021} & tr &  0.11 & 0.05 & 0.10 & 0.005 \\
BD-CGZSL~\cite{kuchibhotla2022unseen} & tr & 0.18 & 0.14 & 0.13 & -0.02  \\ \midrule
CN-ZSL~\cite{skorokhodov2021class} & in & -  & - & -0.04 & -0.02  \\
BD-CGZSL-in~\cite{kuchibhotla2022unseen} & in & \textbf{0.18}  & \textbf{0.15} & 0.14 & -0.03 \\ \midrule
\textbf{ours} + interpolation & in & 0.12 & 0.10 & \textbf{0.19} & \textbf{0.01} \\
\textbf{ours} + dictionary & in & 0.11 & 0.11 & \textbf{0.19} & \textbf{0.01} \\ \bottomrule
\end{tabular}
\caption{Backward transfer of different CGZSL methods, where higher results indicate less forgetting. 
}\label{tab:forgetting}
\end{table}
Table \ref{tab:forgetting} displays the backward transfer of different continual zero-shot algorithms, where higher results indicate better knowledge retention. Our model exhibits a strong backward transfer capability, particularly on longer task sequences where it is needed the most. We achieved the highest BWT score of 0.19 on CUB. On SUN, negative BWT scores (i.e., forgetting) are observed in most other models, but our method can still retain knowledge from the past. These results suggest that the analysis tools we created allow us to identify the critical factors for zero-shot learning, and the development of tools for continual learning can improve our ability to retain information.

\subsection{Ablation Study}
\begin{table}[b]\centering \small
\begin{tabular}{lcc}
\toprule
 & \multicolumn{1}{c}{interpolation} & \multicolumn{1}{c}{dictionary} \\ \midrule
with ${R}_\text{GRW}$ + $L_\text{GRW}$ & \textbf{28.4} & \textbf{28.8} \\ 
 $\,\,\,\,\,\,\,\,$ -  $L_\text{creativity}$  & 27.72 & 27.66 \\ \midrule 
w/o ${R}_\text{GRW}$, $L_\text{GRW}$ & 19.07  & 20.75  \\ 
 $\,\,\,\,\,\,\,\,$ -  $L_\text{creativity}$ & 14.43 & 14.43 \\  \midrule 
with $L_\text{GRW}$  & 26.73 & 27.39 \\ \bottomrule
\end{tabular}
\caption{Effect of the random walk-based penalty with mH measure on CUB dataset.}\label{tab:rw}
\end{table}
To assess the impact of our novel random walk-based penalties, $L_{GRW}$ and $\mathcal{R}_{GRW}$, we conducted ablation experiments; see Table \ref{tab:rw}. he results in Table \ref{tab:rw} indicate that the improvements are mainly attributed to $L_{GRW}$, while $\mathcal{R}_{GRW}$ contributes an additional $1\%$. $L_\text{creativity}$ is also part of the inductive loss. Additionally, removing $L\text{creativity}$ while using our GRW losses has little effect on the performance, as shown in Table \ref{tab:rw}. More details can be found in Appendix \ref{appen:czsl}.

\subsection{More CZSL settings}
Our focus lies on assessing performance under varying seen/unseen class ratios during knowledge accumulation, which is proposed in ~\cite{skorokhodov2021class} and referred to as static setting in ~\cite{kuchibhotla2022unseen}. There are other continual zero-shot learning settings   proposed in ~\cite{kuchibhotla2022unseen}, such as dynamic and online settings. In the dynamic setting, the seen and unseen classes dynamically increase, while in the online setting, certain unseen classes are continually converted to seen classes.  We find our explored static setting a more informing benchmark for the inductive CZSL skill, as the evaluation after every task is always performed on all classes in the dataset and hence is more challenging. Despite this, we still provide a comparison between our method and the baseline methods on the dynamic and online settings; see  Table~\ref{tab:othersettings}. The results show that our method is superior to the baseline in the dynamic and online settings in almost all datasets, and gains the most improvements in the most challenging static setting. 
\begin{table}[t]
\centering  
\footnotesize\centering
\begin{tabular}{llllll}
\toprule
 & setting & {AWA1} & {AWA2} & {CUB} & {SUN} \\\midrule
BD-CGZSL & D & 56.9/49.1  & 56.4 & 16.8 & 28.0 \\
\textbf{ours + inter.} & D  &\textbf{60.0}& \textbf{58.8}& \textbf{32.8} &\textbf{ 41.6} \\
\textbf{ours + dic.} & D  & 59.7 & 55.5 & 31.8 &  40.2\\ \midrule
BD-CGZSL & O & 56.9/49.1  & \textbf{53.4} & 28.4 & 33.7 \\
\textbf{ours + inter.} & O  &\textbf{49.6}& 48.5& \textbf{32.3} &\textbf{39.6} \\
\textbf{ours + dic.} & O  & 46.1 & 47.3 & 31.2 &  39.2\\ 
\bottomrule
\end{tabular}\caption{mH in dynamic setting (D) and online setting(O) }\label{tab:othersettings}
\end{table}

\subsection{Replay Buffer Analysis}
Some existing methods~\cite{ghosh_adversarial_2021, ghosh_dynamic_2021, kuchibhotla2022unseen} tend to use the generative replay method proposed by \cite{gautam_generative_2021}, where the correctly predicted seen generated features from the previous task are stored in buffers. However, the buffer size increases significantly over tasks since a fixed number of samples for each class is stored, and if the model struggles to make accurate predictions for certain classes, samples from these classes are absent in the buffer.

We  empirically found that the class-balanced experience replay method proposed by \cite{prabhu2020gdumb} can be extremely helpful. At every task, we save the class attribute in $\mA^{1:t}$, the class center matrix $\mC$, and modify the buffer with current features noted as $\mD_s^{1:t}$, such that the buffer is balanced across all the seen classes. 
\begin{table}[t]\centering\footnotesize
\begin{tabular}{ccccccc}
\toprule
 & \multirow{2}{*}{\begin{tabular}[c]{@{}c@{}} Buffer \\Size\end{tabular}} &  & \multicolumn{2}{c}{Ours} &  & BD-CGZSL (\textit{tr)} \\ \cmidrule{4-5} \cmidrule{7-7} 
 &  &  & BWT & mHA &  & mHA \\ \midrule
generative & 28.5k &  & 0.14 & 21.06 &  & 17.76 \\
real & 10k &  & 0.17 & 28.44 &  & 27.79 \\
real  & 5k &  & 0.19 & 28.8 &  & 26.55 \\
real & 2.5k &  & 0.08 & 26.99 &  & 26.77 \\ \bottomrule
\end{tabular}
\caption{Comparison of generative replay and real replay methods on CUB~\cite{WahCUB_200_2011}. Dictionary-based attribute generation is used}\label{tab:replay}
\end{table}

In this comparison on the CUB dataset, we observe in Table \ref{tab:replay} that the method using real replay can achieve better harmonic accuracy with a smaller buffer size (around 1/10 of the generative replay buffer size) and comparable backward transfer with a slightly larger buffer size (around 1/5 of the generative replay buffer size). Moreover, the real replay-based method is not as sensitive to the buffer size as the generative replay-based methods. It is worth noting that DVGR, A-CGZSL, and BD-CGZSL typically use generative replay, while only CN-CGZSL uses real replay. In addition, the last column in Table \ref{tab:replay} shows that our proposed real replay method can also improve the harmonic accuracy of other methods. 

\begin{figure}
    \centering
\includegraphics[width=\linewidth]{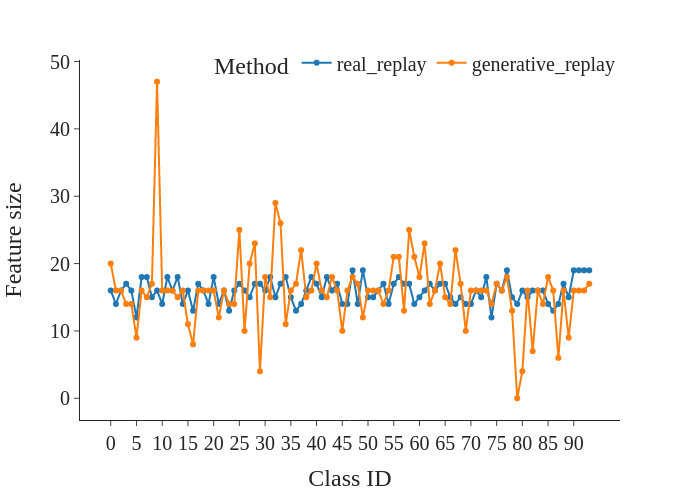}
\caption{Comparison of replayed number of features per class in different replay method at task 2 in SUN dataset}\label{fig:replay}
\end{figure}
To understand the real replay and generative replay, we extend our analysis by visualizing the distribution of buffer features across various classes in task 2 of the SUN dataset, as illustrated in Figure \ref{fig:replay}. Real replay approach exhibits a balanced allocation of features across all classes. Conversely, the generative replay technique displays an intriguing pattern, wherein certain classes lack a substantial number of stored features, while others exhibit a twofold increase. Notably, the classes with fewer stored features coincide with instances where the model's performance is suboptimal. This discrepancy can be attributed to the generative replay method's propensity to store exclusively the accurately classified generated data. Consequently, this uneven distribution of stored features could potentially lead to a compromised performance in these classes during subsequent tasks.

\section{Conclusion and Discussion}

In this paper, we focus on inductive continual zero-shot learning (CZSL) to eliminate the need of unseen information for more realistic learning systems. To this end, we developed a framework for the theoretical analysis of generative zero-shot learning,  introducing a distance metric to measure the ability of generated samples to represent the unseen space when the unseen information is inaccessible during training. We also proposed a continual zero-shot algorithm, ICGZSL, which can reduce the distance without using unseen information during training. We conducted experiments on four popular continual zero-shot learning benchmarks: AWA1, AWA2, CUB, and SUN. Our approach achieved around $3\%$ higher harmonic accuracy in the small dataset and around $7\%$ in the larger dataset compared to previous inductive and transductive methods. These results demonstrate that unseen semantic information is not essential when a well-analyzed seen distribution and method are used.

However, it is important to acknowledge certain limitations in our work. While the developed theoretical bounds and distance measures hold promise for methodical numeric analysis, a more stringent alignment between empirical and anticipated distance measures could substantially enhance algorithmic design. Moreover, the consideration of multi-class classification conditions warrants attention. Additionally, the use of a frozen backbone for image feature extraction, though effective, encourages further exploration into continual learning methods that facilitate viable zero-shot learning capabilities while enabling the backbone to progressively accumulate knowledge.

~\clearpage
{
\bibliographystyle{ieee_fullname}
\bibliography{refer}
}
\appendix
\onecolumn
\section{Derivation of Theorems in Section ~\ref{sec:bound}}
\subsection{Theorem ~\ref{thm:gbzsl}} \label{appen:thm1}

\textbf{Notation Statement in the Appendix.} To involve risk between hypotheses and between hypothesis and ground truth models, we use $\epsilon_\cdot(h,f)$ or $\epsilon_\cdot(h,h^*)$ to specify which space the risk is computed on.

\begin{defn}[$\mathcal{H}\Delta \mathcal{H}$-distance~\cite{ben-david_theory_2010}]
Given two feature distributions $\mathcal{D}_g$ and $\mathcal{D}_r$, and the hypothesis class $\mathcal{H}$, the $\mathcal{H}\Delta \mathcal{H}$-distance between $\mathcal{D}_g$ and $\mathcal{D}_r$ is defined as
\begin{equation}
    \begin{split}
        d_{\mathcal{H}\Delta \mathcal{H}}(\mathcal{D}_g , \mathcal{D}_r ) = 2\sup_{h,h' \in \mathcal{H}}|&\Prob_{\vx \sim \mathcal{D}_g} [h(\vx) \neq h' (\vx)] - \Prob_{\vx \sim \mathcal{D}_r}   h(\vx) \neq h'(\vx)| \enspace.
    \end{split}
\end{equation}
\end{defn}

We first define the expected version of Definition \ref{defn:1}
\begin{defn}[Expected Generative distance]
\label{defn:dhappen}
Given two feature distributions $\mathcal{D}_{h}$ and $\mathcal{D}_u$, the ground truth labelling function $f_{h}, f_u$, and the optimal hypothesis $h^* = \argmin_{h\in H}\epsilon(h,f_{h}) + \epsilon(h,f_s)$ of a model training on the distribution $\gD_s, \gD_{h}$. The $h^*\Delta f$-distance between $\mathcal{D}_{h}$ and $\mathcal{D}_u$ is defined as
\begin{equation}
    \begin{split}
    d_{h^*}(\mathcal{D}_{h},\mathcal{D}_u) = & |\Prob_{(\vx,\va) \sim \mathcal{D}_{h}}[f_{h}(\vx) \neq h^* (\vx,\va)] \\
    &  - \Prob_{(\vx,\va) \sim \mathcal{D}_u}  [ f_u(\vx) \neq h^* (\vx,\va)]| \enspace,
\end{split}
\end{equation}
\end{defn}

Note that the following inequality related to $d_{\mathcal{H}\Delta \mathcal{H}}(\mathcal{D}_g , \mathcal{D}_r ) $ holds for any $h$ and $h^*$
\begin{equation} \label{eq:rewritedhh}
\begin{split}
    d_{\mathcal{H}\Delta \mathcal{H}}(\mathcal{D}_g , \mathcal{D}_r ) & =2\sup_{h,h' \in \mathcal{H}}|\Prob_{\vx \sim \mathcal{D}_g} [h(\vx) \neq h' (\vx)]  - \Prob_{\vx \sim \mathcal{D}_r}   h(\vx) \neq h'(\vx)|\\
    & \geq 2|\Expect_{\vx \sim \mathcal{D}_g}[\ones_{h(\vx) \neq h^* (\vx)}] - \Expect_{\vx \sim \mathcal{D}_r} [ \ones_{ h(\vx) \neq h^* (\vx)}]|\\
    & = 2|\epsilon_g(h,h^*) - \epsilon_r(h,h^*)| \enspace.
\end{split}
\end{equation}

\begin{lem}[\cite{abu2012learning}]\label{lem:actualrisk}
For a fixed hypothesis, the actual risk can be estimated from the empirical error with probability $1-\delta$
\begin{equation}
    \epsilon(h,f) \leq \hat{\epsilon}(h,f) + \sqrt{\frac{1}{2m}\log \frac{2}{\delta}} \enspace,
\end{equation}
where $\epsilon(h,f)$ is the actual risk, $\hat{\epsilon}(h,f)$ is the empirical risk, and $m$ is the number of testing samples.
\end{lem}

\begin{prop}[Bound $d_{h^*}(\mathcal{D}_u,\mathcal{D}_{h})$ by $\bar{d}_{GDB}(\mD_u, \mD_{h})$ ]\label{prop:GDBrelation}
The distribution distance $d_{h^*}(\mathcal{D}_u,\mathcal{D}_{h})$ can be bounded by it empirical counterpart by 
\begin{equation}
    d_{h^*}(\mathcal{D}_u,\mathcal{D}_{h}) \leq \bar{d}_{GDB}(\mD_u, \mD_{h}) + C(\frac{1}{m},\frac{1}{\delta}) \enspace,
\end{equation}
where $C(\frac{1}{m},\frac{1}{\delta})$ is a constant term depending on the training sample size $m$ and confidence $1-\delta$. Here $\mathcal{D}_\cdot$ represent the distribution, and $\mD_\cdot$ represents the dataset sampled from the corresponding distribution.
\begin{proof}
Similar to Equation \ref{eq:rewritedhh}, we can write our generative distance as 
\begin{equation}
    d_{h^*}(\mathcal{D}_u,\mathcal{D}_{h}) = 2|\epsilon_{h}(h^*,f) - \epsilon_u(h^*,f)| \enspace.
\end{equation}
Combining Lemma \ref{lem:actualrisk}, we have 
\begin{equation} \label{eq:GDBrelation}
\begin{split}
     \frac{1}{2} d_{h^*}(\mathcal{D}_u,\mathcal{D}_{h}) &= |\epsilon_{h}(h^*,f) - \epsilon_u(h^*,f)|\\
    & \leq |\hat{\epsilon}_{h}({h}^*,f) - \hat{\epsilon}_u({h}^*,f)|+ |(\hat{\epsilon}_{h}({h}^*,f) + \hat{\epsilon}_u(h^*,f)) - (\epsilon_{h}(h^*,f)  + \epsilon_u(h^*,f))| \\
    & \lesssim \frac{1}{2} \bar{d}_{GDB}(\mD_u, \mD_{h}) + C(\frac{1}{m},\frac{1}{\delta}) \enspace,
\end{split}
\end{equation}
where $h^* = \argmin_{h'\in H}\epsilon_s(h',f_s) + \epsilon_{h}(h',f_{h})$, and $\hat{h}^* = \argmin_{h'\in H}\hat{\epsilon}_s(h',f_s) + \hat{\epsilon}_{h}(h',f_{h})$. Following the discussion of  \cite{chang_g2r_2019}, we assume the optimal hypothesis $\hat{h}^*$ we can achieve is very close to the global minimum when the training sample is large, then we can estimate $h^*$ in Equation \ref{eq:GDBrelation} by $\hat{h}^*$. $C(\frac{1}{m},\frac{1}{\delta})$ is obtained from Lemma \ref{lem:actualrisk}
\end{proof}
\end{prop}

\paragraph{Proof of theorem ~\ref{thm:gbzsl}}
Given the CZSL procedure described in section \ref{sec:problemsetup}, with confidence $1-\delta$ the risk on the unseen distribution is bounded by 
\begin{equation} \label{eq:main}\footnotesize
\begin{split}
    \epsilon(h, f_u^t) \leq  \hat{\epsilon}(\hat{h}^*,f_{s}^{1:t})  + \frac{1}{2}d_{\mathcal{H}\Delta \mathcal{H}}(\mathcal{D}_{s}^{1:t},\mathcal{D}^t_u)+ \bar{\lambda} 
    + \frac{1}{2} \bar{d}_{GDB}(\mD^t_u, \mD^t_{h}) 
\end{split}
\end{equation}
where $\hat{h}^* = \argmin_{h\in H}\sum_{i=1}^t\hat{\epsilon}(h,f_{s}^i) + \hat{\epsilon}(h,f^t_{h})$, $\bar{\lambda} = \hat{\epsilon}(\hat{h}^*,f_{s}^{1:t}) + \hat{\epsilon}(\hat{h}^*,f^t_{h})$. 

\begin{proof}
Let $h^* = \argmin_{h\in H}\sum_{i=1}^t{\epsilon}(h,f_{s}^i) + {\epsilon}(h,f^t_{h})$, and $\lambda = {\epsilon}({h}^*,f_{s}^{1:t}) + {\epsilon}({h}^*,f^t_{h})$. We write $\epsilon_s(\cdot,\cdot)$ as the union seen distribution from time $1:t$. Then
\begin{equation} \label{eq:thm1}
\begin{split}
    &\epsilon_u(h,f_u) \\
    &= \epsilon_s(h,f_s)  + \epsilon_u(h,h^*) - \epsilon_s(h,h^*) + \epsilon_{h}(h^*,f_{h}) + \epsilon_s(h^*,f_s) - \epsilon_{h}(h^*,f) + \epsilon_u(h^*,f)\\
    & - \epsilon_s(h,f_s) - \epsilon_u(h,h^*) + \epsilon_s(h,h^*)  -  \epsilon_s(h^*,f_s) -  \epsilon_u(h^*,f)+ \epsilon_u(h,f_u) \\
    & \leq \epsilon_s(h,f_s)  + | \epsilon_u(h,h^*) - \epsilon_s(h,h^*)| + |\epsilon_{h}(h^*,f_{h}) + \epsilon_s(h^*,f_s)| + | \epsilon_{h}(h^*,f) - \epsilon_u(h^*,f) |\\
    & - \epsilon_s(h,f_s) + \epsilon_s(h,h^*) - \epsilon_s(h^*,f_s) - \epsilon_u(h,h^*) + \epsilon_u(h,f_u) -\epsilon_u(h^*,f) \\
    & \leq \epsilon_s(h,f_s)  + d_{\mathcal{H}\Delta \mathcal{H}}(\mathcal{D}_{s}^{1:t},\mathcal{D}^t_u) + \lambda+ d_{h^*}(\mathcal{D}_{h},\mathcal{D}_u)  - \epsilon_s(h,f_s) + \epsilon_s(h,h^*) \\
    & - \epsilon_s(h^*,f_s) - \epsilon_u(h,h^*) + \epsilon_u(h,f_u) -\epsilon_u(h^*,f) \enspace.
\end{split}
\end{equation}
Note that for any distribution
\begin{equation} \label{eq:thm1triangle}
\begin{split}
    |\epsilon_\mathcal{D}(h,f_\mathcal{D}) - \epsilon_\mathcal{D}(h,h^*)|& = |\Expect_{\vx\sim \mathcal{D}}[\ones_{h\ne f_\mathcal{D}}] - \Expect_{\vx\sim \mathcal{D}}[\ones_{h\ne h^*}]| \\
    & = |\Expect_{\vx\sim \mathcal{D}}[\ones_{h\ne f_\mathcal{D}} - \ones_{h\ne h^*}]| \\
    & \leq  \Expect_{\vx\sim \mathcal{D}}[\ones_{h^*\ne f_\mathcal{D}}] = \epsilon_\mathcal{D}(h^*,f_\mathcal{D}) \enspace,
\end{split}
\end{equation}
where the inequality holds by the triangle inequality of the characteristic function, \ie, $\ones[a\ne b] \geq \ones[a \ne c] - \ones[b \ne c]$ for $\forall a,b,c \in \reals$. Equation (\ref{eq:thm1triangle}) shows that the fourth line in Equation (\ref{eq:thm1}) is less than or equal to zero.

Combining Equation \ref{eq:thm1triangle}, the Equation \ref{eq:thm1} can be written as
\begin{equation} \label{eq:expectthm1}
\begin{split}
    \epsilon_u(h,f_u) \leq &\epsilon_s(h,f_s)  + \frac{1}{2}d_{\mathcal{H}\Delta \mathcal{H}}(\mathcal{D}_{s}^{1:t},\mathcal{D}^t_u) + \lambda + d_{h^*}(\mathcal{D}_{h},\mathcal{D}_u)\enspace,
\end{split}
\end{equation}
However, Equation (\ref{eq:expectthm1}) involves unknown risk and unsolvable distribution. We combine the expected risk and the actual observed risk by Lemma \ref{lem:actualrisk}. Let $h^* = \argmin_{h\in H}\sum_{i=1}^t\hat{\epsilon}(h,f_{s}^i) + \hat{\epsilon}(h,f^t_{h})$ be the optimal hypothesis on the training set, and $\bar{\lambda} = \hat{\epsilon}(\hat{h}^*,f_{s}^{1:t}) + \hat{\epsilon}(\hat{h}^*,f^t_{h})$, we have  $\lambda \leq \bar{\lambda}$.  Together with Lemma \ref{lem:actualrisk} and Proposition \ref{prop:GDBrelation}, we have 
\begin{equation} \label{appen:eqfinal}
\begin{split}
    \epsilon_u \leq &\sum_{i=1}^t\alpha^i(\hat{\epsilon}(h,f_{s^i})  + \frac{1}{2}d_{\mathcal{H}\Delta \mathcal{H}}(\mathcal{D}_{s^i},\mathcal{D}_u)) + \bar{\lambda} + \frac{1}{2} \bar{d}_{GDB}(\mD_u, \mD_{h}) + C (\frac{1}{m} + \frac{1}{\delta}) \enspace,
\end{split}
\end{equation}
\end{proof}

\subsection{Explanation of Statement ~\ref{stat:main}} \label{appen:stat}
Let $\mD_{h} \sim \mathcal{D}_{h}$ be the generated unseen set we are training on, where $\mathcal{D}_{h}$ is the empirical distribution of all possible generations. In unsupervised domain adaptation, \cite{van2017unsupervised} uses random walk to select label set for the samples who have small generalization error. Proposition 3.2 of \cite{van2017unsupervised} demonstrates that the self transition probability of a Markov chain represents an upper bound on the margin linear classifier's generalization error. This concept is adapted to connect our GDB bound connected to the Markov Chain in below. In our sample generation procedure, we generate only one sample from each class. Our discussion of this section will be based on this.  We have $\bar{d}_{GDB}(\mD_u, \mD_{h}) \propto -\sum_{i \in I_u} \Prob(\va_{u[i]} \in \mD_{h})$, where the probability is taken over $\mathcal{D}_{h}$, and $I_u$ is the index set of unseen real attributes. This is because the difference of the risk will be reduced if the generations contain as many points close to ground-truth  unseen ones as possible. Consider the Markov chain with single step transition probabilities $p_{ij}$ of jumping from node $i$ to node $j$. Each node represents a generated sample. Let 
\begin{equation}
    p_{ij} = \Prob[h(\vx_i)=y_j] \enspace,
\end{equation}
where $h$ is the hypothesis trained on $\mD_{h}$, and the $h$ output predictions on the current generation's classification space depending on the quality of $h$, and the probability is taken over $\mathcal{D}_{h}$. We assume the  training achieves error $\epsilon$, then $h(\vx_i) = y_i$ with probability $(1 - \delta)$  if the training set contains class with attribute $\va_i$.  It is not hard to prove that $\Prob(\va_{u[i]} \in \mD_{h})  \geq  p_{ii}(1 - \delta)(1-\epsilon)$ by the generalization bound, since if $\va_{u[i]} \notin \mD_{h}$, $y_{u[i]}$ is not in the current generation's classification space.  It follows that 
\begin{equation}
    \begin{split}
        \bar{d}_{GDB}(\mD_u, \mD_{h}) \propto &- \sum_i\Prob(\va_{u[i]}\subseteq \mD_{h}) \leq -\sum_i p_{ii}(1 - \delta)(1-\epsilon)
    \end{split}
\end{equation}

Then we can release the bound $\bar{d}_{GDB}(\mD_u, \mD_{h})$  by increasing $\sum_i p_{ii}$. Note that $\Prob(\va_{u[i]} \in \mD_{h})$ can be replaced by $\Prob(\min_{\va_{h[j]}\in\mD_{h}}|\va_{u[i]}- \va_{h[j]}|<\varepsilon)$ with the robustness assumption of the model. 

When two generations have the same $\sum_ip_{ii}$, we prefer the one having higher diversity. The diversity of the generated set $\mD_{h}$ can be quantified from the perspective of determinantal point process. As mentioned in \cite{kang2013fast} and \cite{elfeki2019gdpp}, Determinantal Point Process (DPP)  is a framework for representing a probability distribution that models diversity. More specifically, a DPP over the set $\mathcal{V}$ with $|\mathcal{V}| = N$, given a positive-definite similarity matrix $L \in \reals^{N\times N}$, is a probability distribution $P_L$ over any $S \subseteq \mathcal{V}$ in the following form
\begin{equation}
   P_L[S] \varpropto \det (L_s) \enspace,  
\end{equation}

where $L_s$ is the similarity kernel of the subset $S$\footnote{The feature representation of the similarity space is typically normalized so the highest eigen value is 1, and hence the determinant (multiplication of the eigen values) is $<$ 1}.
Since the point process according to this probability distribution naturally capture the notion of diversity, we hope to generate a subset with high $P_L[\mD_{h}]$ where the $\mathcal{V}$ is viewed as $\mathcal{D}_{h}$  and the transition matrix is viewed as the similarity kernel.  One way to generate a set of unseen samples with high $\det (L_{\mD_{h}})$ is to encourage the diagonality of the transition matrix, which can be achieved by promoting orthogonality of the generated samples. Moreover, since actually $f_{h}$ is a look-up table, low $\sum_{j\ne i} p_{ji}$ can be explained as the large dis-similarity of the generated unseen samples from different class.

\section{More Details of Section ~\ref{sec:method}}\label{appen:method}
Algorithm \ref{alg:main} shows the overall training process. The Discriminator and Generator are alternatively optimized. During the training of the Generator (line 11 – 22), we propose to generate unseen attributes (line 12 for interpolation-based method and line 12,13 for dictionary-based method) and encourage the generations to be realistic and deviate from the seen generations (line 19). After the training of each task, we propose to store the current semantic information and real features in the buffer.

\subsection{Regularization terms in Loss Function 4} \label{appen:fullalg}
We closely follow \cite{kuchibhotla2022unseen} for the regularization terms of the Generator and Discriminator. The regularization term on discriminator encourages the semantic embedding to be close to the class center, i.e., at task $t$
\begin{equation}
  \mathcal{R}^t_D = \|D(\mA^{1:t}_s) - {\mC}^{1:t}_s\|^2_{\mathsf{F}} \enspace,  
\end{equation}
where $\mA^{1:t}_s$ is the attribute matrix and ${\mC}^{1:t}_s$ is the class mean matrix computed by seen features up to the current task. $\|\cdot\|_{\mathsf{F}} $ is the Frobenius norm. The regularization terms on the generator encourage the seen generations to be close to the seen class centers and have moderately distanced to their semantic neighborhoods. $\mathcal{R}_G$ is defined as
\begin{equation}
    \mathcal{R}_G = L_\text{nuclear} + L_\text{sal} \enspace.
\end{equation}
$ L_\text{nuclear}$ is the Nuclear loss, defined as
\begin{equation}
    L_\text{nuclear} =  \|{\mC}^t_s - {\mC}^t_{sg}\|^2_{\mathsf{F}} \enspace,
\end{equation}
where ${\mC}^t_s$ is the class mean matrix computed by seen features of current task, and ${\mC}^t_{sg}$ is the class mean matrix computed by generated seen features of current task. $L_\text{sal}$ is the incremental bidirectional semantic alignment loss defined as
\begin{equation}
    \begin{split}
    L_\text{sal} = &\frac{1}{N_s^t} \sum_{i=1}^{N_s^t} \sum_{j \in \mI_i}\|  
    \max \{0,\langle {\mC}_{s[j]} , {\mC}_{sg[i]} \rangle - (\langle \mA^t_{s[i]}, \mA^t_{s[j]}  \rangle + \varepsilon)\} \|^2 \\
    &  + \|\max \{0, (\langle \mA^t_{s[i]}, \mA^t_{s[j]}\rangle - \varepsilon) - \langle {\mC}_{s[j]}, {\mC}_{sg[i]} \rangle\} \|^2 \enspace,
\end{split}
\end{equation}
where $N_s^t$ is the number of current seen classes at task $t$, $\mI_i$ is the neighbor set of class $i$, $\varepsilon$ is the margin error, $\langle \cdot, \cdot \rangle$ is the cosine similarity.

\subsection{Visualization of Attribute Distribution} \label{appen:mehtodvisual}
\begin{figure}
\includegraphics[width=.48\textwidth]{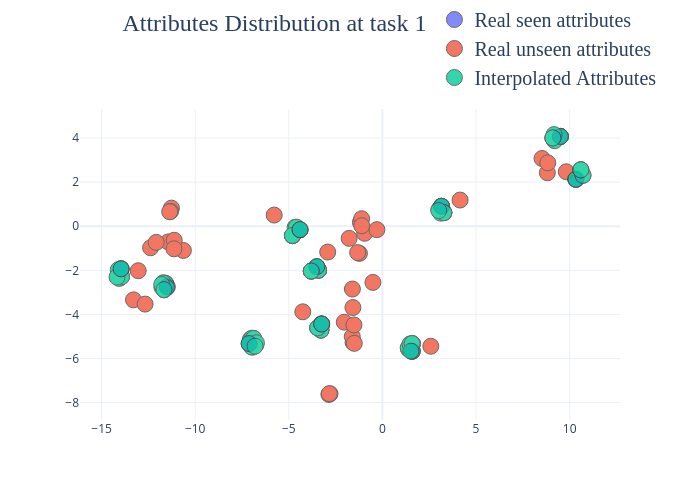}
\includegraphics[width=.48\textwidth]{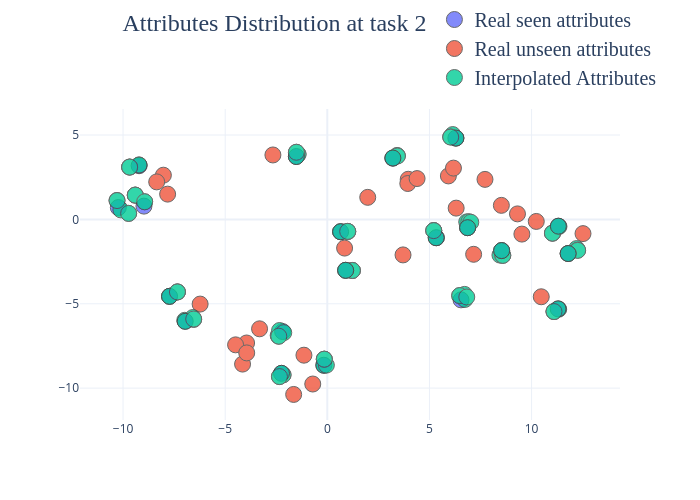}

\includegraphics[width=.48\textwidth]{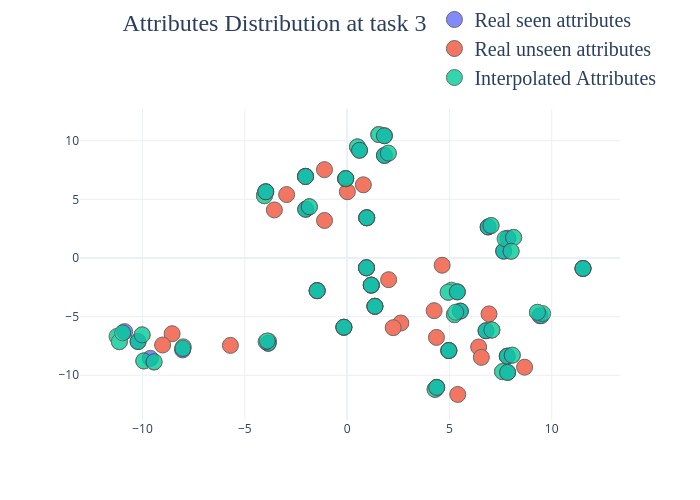}
\includegraphics[width=.48\textwidth]{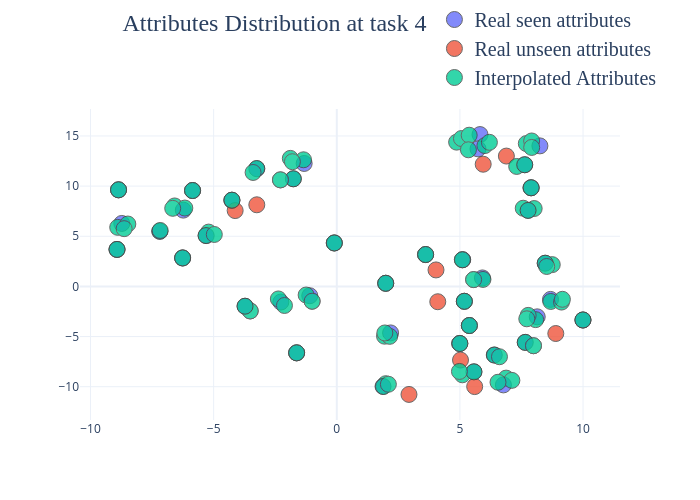}
\caption{Attribute distribution T-SNE visualizations of AWA1 dataset in different task with interpolation method}\label{appen:attdist}
\end{figure}

In our analysis, we assume that the hallucinated attributes can effectively represent the real unseen attributes compactly. To visualize the distribution of these attributes, we employ the T-SNE embedding method. As shown in Figure \ref{appen:attdist}, the plot illustrates the distribution of seen attributes, unseen attributes, and hallucinated attributes across different tasks. It is important to note that only a partial subset of the hallucinated attributes for each task is displayed in the plot, while the actual number of hallucinated attributes is equivalent to the number of training samples.

As the task progresses and the learner is exposed to more seen classes, the hallucinated attributes become more aligned with the unseen attribute. However, in areas where the distribution of unseen attributes is sparse (as indicated by the blank regions in Figure \ref{appen:attdist}), the hallucinated attributes are also sparse. In such cases, the hallucinated attributes tend to describe the potential visual space, deviating from the seen attributes, and providing compact support for the unseen attributes. This aligns with our assumptions and demonstrates the efficacy of our approach.

\subsection{Numerical Verification of GRW Loss} \label{appen:numericalgrw}
\begin{figure}
\centering
  \includegraphics[width=.5\textwidth]{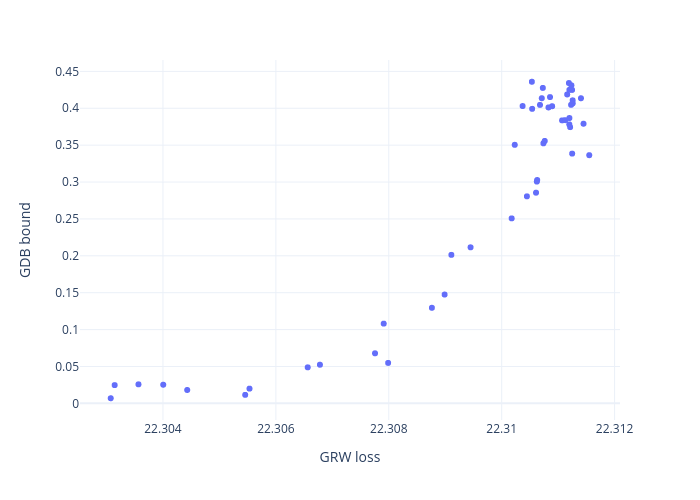}
  \caption{Relationship between $\bar{d}_{GDB}$ and the GRW loss in CUB dataset}\label{fig:loss-bound}
\end{figure}

In statement 3.3, we asserted that the GRW loss can effectively reduce $\bar{d}_{GDB}$. To demonstrate the relationship between the GRW loss and the bound $\bar{d}_{GDB}$, we plotted a figure using the model at different epochs for different $\hat{h}^*$. In this figure, we used the difference between the generated hallucinated samples accuracy and the test unseen accuracy to represent $\bar{d}_{GDB} = |\hat{\epsilon}_u - \hat{\epsilon}_{h}|$ at a randomly selected task. As shown in Figure \ref{fig:loss-bound}, we observed a strong positive correlation between the GRW loss and $\bar{d}_{GDB}$, particularly, when the loss decreases. This finding suggests that by minimizing the GRW loss, we can reduce the bound between the generated hallucinated space and the true unseen space.


\subsection{Relation to Other Work using Random Walk} \label{appensec:relation}
We adapt random walk modeling~\cite{ayyad2020semi} with three key changes. 
\begin{enumerate}
    \item Previous works such as \cite{ayyad2020semi, 8099557} have represented class prototypes or centers using a few examples provided for each class. However, in our setting, we aim to deviate from seen classes and facilitate knowledge transfer to unseen classes through attributes or semantic descriptions. To achieve this, we define the seen class centers $\mC$ in a semantically guided way by computing the mean of generated seen samples from their corresponding attributes. Specifically, we define $[\mC]_i$ as the mean of generated samples from the attribute vector $a_i$ for class $i$, i.e., $[\mC]_i = \text{mean } G( z,a_i)$, where $G$ is the generator and $z$ is the noise vector.
    \item  \cite{ayyad2020semi, 8099557} use unlabeled data points to calculate the random walk, where we use generated examples.
    \item In contrast to the few-shot learning problem where class prototypes are computed using unlabeled examples of seen classes, our approach generates examples from hallucinated classes. Thus, the loss functions proposed in \cite{ayyad2020semi, 8099557} aim to attract unlabeled samples to labeled samples, whereas our goal is to push hallucinated samples away from seen samples. In \cite{8099557}, global consistency is encouraged using a random walk from labeled data to unlabeled data (represented by their $\mA$ matrix) and back to labeled samples (represented by their $\mA^\top$ matrix). The aim is to promote the identity distribution of paths, where the starting and ending points are of the same class. \cite{ayyad2020semi} investigates a more general case where the number of random walk steps between unlabeled classes is greater than one (represented by their $\mB$ matrix). In our case, as none of the generated hallucinated samples belong to seen classes, we use the random walk approach to encourage uniform distribution instead of identity distribution for all the paths from seen to generated examples of hallucinated classes and back to seen classes, represented by our $\mP^{\mC_s X_h} \mP^{X_hX_h}\mP^{ X_h C_s }$ matrix. This approach provides a deviation signal that encourages the model to learn distinct representations for seen and hallucinated classes, facilitating better knowledge transfer to hallucinated classes.
\end{enumerate}

\cite{van2017unsupervised} focuses on unsupervised domain adaptation, which involves doing a random walk over all potential labeling circumstances on unlabeled target data to identify a stationary labeling distribution. Labeling stability is defined from the perspective of a generalization bound which can be attained through a stationary Markov chain. We borrow the idea of using the Markov chain to estimate the relationship between different labeling to find a stationary one that can reduce the generalization bound. We employ the Markov chain to estimate the relationship between different hallucinated generations and discover a diverse one that can reduce the generalization bound. The $L_{GRW}$ loss encourages the random walk to find a highly diverse hallucinated generation, which in turn reduces the generalization bound.

\RestyleAlgo{ruled}

\begin{algorithm*}
  \SetKwData{Left}{left}\SetKwData{This}{this}\SetKwData{IN}{in}
  \SetKwFunction{Union}{Union}\SetKwFunction{FindCompress}{FindCompress}
  \SetKwInOut{Input}{Input}
  \SetKwInOut{Output}{output}
  \SetKwInOut{Data}{Data}
  \SetKwInOut{Init}{Initialize}
  \Input{Total task number $T$, training epoch $E$, random walk length $R$, decay rate of random walk $\gamma$, and coefficients $\lambda_{c,rd,i,rg}$, learning rate $\alpha_{G,D,Dic}$, buffer size $B$}
  \Data{$X^{1:T}_{s}$, $y^{1:T}_{s}$, $a^{1:T}_{s}$}
  \Init{Generator, Discriminator}
    \BlankLine
  \For{ $t = 1:T$}{
    Get train loader by concatenating train set $t$ with buffer data\;
    \For{$e=1:E$}{
        Get $X_s^t, y_s^t$ sampled from train loader. Get $a_s^{1:t}$ from current train set and buffer \; 
        \Begin( Train Discriminator){
        Generate samples conditioning on seen attributes $X^t_{sg} = G(z,a_s^t)$ \;
        Compute real-fake loss $\mathcal{L}_\text{real-fake}$ in equation (5) using real seen samples $X_s^t$, generated seen samples $X^t_{sg}$, and current task attribute $a^t_s$\;
        Compute classification loss $\mathcal{L}_\text{classification}$ in equation 6 using real seen samples $X_s^t$, generated seen samples $X^t_{sg}$, and attributes $a^{1:t}_s$\;
        Compute $\mathcal{L}_D$ in equation 4 and update $\theta_D \gets \theta_D - \alpha_D \nabla \mathcal{L}_D$ \;
        }
        \Begin( Train Generator){
        Generate $a^{t}_{ug}$ by interpolation between two random $a^t_s$ \;
        \If{Use dictionary based method}{
        Initialize the dictionary with the interpolated attribute and get $\theta_{Dic}$
        }
        Generate samples conditioning on unseen attributes $X^t_{ug} = G(z,a_{ug}^t)$ \;
        Compute the second part of real-fake loss $\mathcal{L}_\text{real-fake}$ in equation (5) using  generated unseen samples $X^t_{sg}$ and current task attribute $a^t_s$\;
        Compute the second part of classification loss $\mathcal{L}_\text{classification}$ in equation 6 using generated unseen samples $X^t_{sg}$ and attributes $a^{1:t}_s$\;
        Compute tehe inductive loss in $\mathcal{L}_\text{inductive}$ using $C_s^t = \text{mean}(X^t_s)$, generated seen samples $X_{sg}^t$, and unseen generated samples $X_{ug}^t$
        Compute $\mathcal{L}_G$ in equation 4 and update $\theta_G \gets \theta_G - \alpha_G  \nabla \mathcal{L}_G$ \;
        \If{Use dictionary based method}{$\theta_{Dic} \gets \theta_{Dic} - \alpha_{Dic} \nabla \mathcal{L}_D$}
        }}
    \Begin( Replay data){
        Save $a^t_s$ to the buffer\;
        Save current real features with size $B/N_s^{1:t}$ per class, reduce previous features to size $B/N_s^{1:t}$
    
    }
  }
  \caption{Training procedure of ICGZSL}
  \label{alg:main}
\end{algorithm*}

\section{Zero-shot learning experiments} \label{appen:zslexp}
\subsection{Text based zero-shot learning experiments}
Text-based ZSL is more challenging because the descriptions are at the class level and are extracted from Wikipedia, which is noisier.

\paragraph{Benchmarks: } To evaluate the efficacy of zero-shot learning (ZSL) with text descriptions as semantic class descriptions, we conducted experiments on two well-known benchmarks, namely Caltech UCSD Birds-2011 (CUB)\cite{WahCUB_200_2011} and North America Birds (NAB)\cite{Horn2015}. While CUB contains 200 classes with 11,788 images, NAB has 1011 classes with 48,562 images. To gauge the generalization capability of class-level text zero-shot recognition, we split the benchmarks into four subsets: CUB Easy, CUB Hard, NAB Easy, and NAB Hard. The hard splits were designed to ensure that the unseen bird classes from super-categories do not overlap with seen classes, following prior work~\cite{chao2016empirical, Elhoseiny_2018_CVPR,elhoseiny2019creativity}.

\paragraph{Baseline and training: } We introduced a novel GRW loss ($L_{GRW} + \mathcal{R}_{GRW}$) into the inductive zero-shot learning method GAZSL~\cite{Elhoseiny_2018_CVPR} and compared its performance with other inductive zero-shot learning methods. We employed the TF-IDF\cite{salton1988term} representation of the input text for the text representation function $\psi(\cdot)$, followed by an FC noise suppression layer. Our experiments were conducted using a random walk length $R=10$, and we found that longer random walk processes yield better performance in the ablation study. Each ZSL experiment was executed on a single NVIDIA P100 GPU.

\paragraph{Evaluation and metrics: } During the test, the visual features of unseen classes are synthesized by the generator conditioned on a given unseen text description $a_u$, i.e. $x_u = G(s_u, z)$. We generate 60 different synthetic unseen visual features for each unseen class and apply a simple nearest neighbor classifier on top of them.   We use two metrics: standard zero-shot recognition with the Top-1 unseen class accuracy and Seen-Unseen Generalized Zero-shot performance with Area under Seen-Unseen curve~\cite{chao2016empirical}. 

\paragraph{Results:} Our proposed approach improves over older methods on all datasets and achieves SOTA on both Easy and SCE(hard) splits, as shown in Table~\ref{tb:nabcub}.
We show improvements in 0.8-1.8\%  Top-1 accuracy and 1-1.8\% in AUC. GAZSL~\cite{Elhoseiny_2018_CVPR} + GRW also has an improvement of around 2\% over other inductive loss (GAZSL~\cite{Elhoseiny_2018_CVPR} + CIZSL~\cite{elhoseiny2019creativity}).
\begin{table*}[]
\centering\footnotesize
\minipage[t]{0.45\textwidth}%
{
\begin{tabular}{@{}lcccc@{}}
\toprule
Metric &  \multicolumn{4}{c}{Seen-Unseen AUC (\%)} \\ \cmidrule(lr){2-5} 
Dataset & \multicolumn{2}{c}{CUB} & \multicolumn{2}{c}{NAB} \\
Split-Mode &  Easy & Hard & Easy & Hard \\ \midrule
ZSLNS~\cite{Qiao2016} & 14.7 & 4.4 & 9.3 & 2.3 \\
SynC$_{fast}$~\cite{changpinyo2016synthesized}  & 13.1 & 4.0 & 2.7 & 3.5 \\
ZSLPP~\cite{Elhoseiny_2017_CVPR}   & 30.4 & 6.1 & 12.6 & 3.5 \\
FeatGen~\cite{xian_feature_2018}  & 34.1 & 7.4 & 21.3 & 5.6 \\ \midrule
LsrGAN (\textit{tr})~\cite{vyas2020leveraging}  & 39.5 & 12.1 & 23.2 & 6.4\\ 
 \quad +\textbf{GRW}& 39.9\color{blue}{$^{+0.4}$} & 13.3\color{blue}{$^{+1.2}$} & 24.5\color{blue}{$^{+1.3}$} & 6.7\color{blue}{$^{+0.3}$}\\ \midrule
GAZSL (\textit{in})~\cite{Elhoseiny_2018_CVPR} & 35.4 & 8.7 & 20.4 & 5.8\\
\quad + {CIZSL}~\cite{elhoseiny2019creativity}   & 39.2 & 11.9 & 24.5 & 6.4 \\ 
\quad + \textbf{GRW}  & \textbf{40.7} \color{blue}{$^{+5.3}$} & \textbf{13.7}\color{blue}{$^{+5.0}$} & \textbf{25.8}\color{blue}{$^{+5.4}$} & \textbf{7.4} \color{blue}{$^{+1.6}$} \\ 
\bottomrule   
\end{tabular}}
\caption{Showing Seen-Unseen AUC results of ZSL experiments on noisy text description-based datasets \textbf{CUB} and \textbf{NAB}(Easy and Hard Splits)}
\label{tb:nabcub}
\endminipage\hfill
\minipage[t]{0.5\textwidth}%
\begin{tabular}{lcccc}\toprule
\multirow{2}{*}{Setting} & \multicolumn{2}{c}{CUB-Easy} & \multicolumn{2}{c}{CUB-Hard} \\ \cmidrule{2-5}
 ~ & Top-1 Acc  & SU-AUC  & Top1-Acc  &SU-AUC \\\midrule
+ GRW ($R$=1) &45.41 &39.62 &13.79 &12.58 \\
+ GRW ($R$=3) &45.11 &39.25 &14.21 &13.22 \\
+ GRW ($R$=5) &45.40 &40.51 &14.00 &13.07 \\
+ GRW ($R$=10) &\textbf{45.43} &\textbf{40.68} &\textbf{15.51} &\textbf{13.70}\\
\bottomrule
\end{tabular}
\caption{Ablation studies on CUB Dataset (text). Each row shows either baseline deviation losses and GRW losses with different length on GAZSL~\cite{Elhoseiny_2018_CVPR}}
\label{tb:ablation_zsl}
\endminipage\hfill
\end{table*}
\begin{table*}[t]
    \centering\footnotesize
\begin{tabular}{@{}cccccccccc@{}}
\toprule
Metric & \multicolumn{4}{c}{Top-1 Accuracy (\%)} &  & \multicolumn{4}{c}{Seen-Unseen AUC (\%)} \\ \cmidrule(lr){2-5} \cmidrule(l){7-10} 
Dataset & \multicolumn{2}{c}{CUB} & \multicolumn{2}{c}{NAB} &  & \multicolumn{2}{c}{CUB} & \multicolumn{2}{c}{NAB} \\
Split-Mode & Easy & Hard & Easy & Hard &  & Easy & Hard & Easy & Hard \\ \midrule
GAZSL~\cite{Elhoseiny_2018_CVPR}  & 43.7 & 10.3 & 35.6 & 8.6 &  & 35.4 & 8.7 & 20.4 & 5.8 \\ \midrule
GAZSL~\cite{Elhoseiny_2018_CVPR} + \text{GRW} & \textbf{45.4} & \textbf{15.5}  & \textbf{38.4}  & 10.1  &  & \textbf{40.7}  & \textbf{13.7} & \textbf{25.8} & \textbf{7.4} \\ \midrule
GAZSL~\cite{Elhoseiny_2018_CVPR} + only $L_{GRW}$  & 45.3 & 14.8 & 38.2 & \textbf{10.3} &  & 40.1 & 12.8 & 25.8 &7.4\\
\bottomrule
\end{tabular}
\caption{Ablation study using Zero-Shot recognition on \textbf{CUB} \& \textbf{NAB} datasets with two split settings. We experiment with and without the $\mathcal{R}_{GRW}$ (second and last row). The first loss is the baseline method.}
\label{tb:nab_cub_ablation}
\end{table*}

\paragraph{GRW Loss for Transductive ZSL:} To better understand how the GRW improves the consistency of generated seen features space and generated unseen features space, we conduct experiments on semantic transductive zero-shot learning settings. The improvements are solely from the GRW loss with the ground truth semantic information. We choose LsrGAN~\cite{vyas2020leveraging} as the baseline model. Our loss can also improve LsrGAN on text-based datasets on most metrics, ranging from 0.3\%-3.6\%. However, as we expected, the improvement in the purely inductive/more realistic setting is more significant.

\paragraph{Ablation:}
Table~\ref{tb:ablation_zsl} shows the results of our ablation study on the random walk length. We find that the longer random walk performs better, giving higher accuracy and AUC scores for both easy and hard splits for CUB Dataset. With a longer random walk process, the model could have a more holistic view of the generated visual representation that enables better deviation of unseen classes from seen classes. 

GRW loss contains two parts, $L_{GRW}$ and $\mathcal{R}_{GRW}$. Table~\ref{tb:nab_cub_ablation} shows the results of our ablation study on the $\mathcal{R}_{GRW}$ in zero-shot learning. We perform experiments both with $\mathcal{R}_{GRW}$ and without $\mathcal{R}_{GRW}$.  Training failed with NaN gradients in 5\% of the times without $\mathcal{R}_{GRW}$ but 0\%  with $\mathcal{R}_{GRW}$; thus, it is important for the training stability.

\subsection{Attribute based zero-shot learning experiments}
\paragraph{Benchmarks:} We perform these experiments on the AwA2~\cite{lampert2009learning}, aPY~\cite{farhadi2009describing}, and SUN~\cite{patterson_sun_2012} datasets. 

\paragraph{Baseline, training, and evaluation:} We perform experiments on the widely used GBU~\cite{gbu} setup, where we use class attributes as semantic descriptors. The evaluation process and training devices are the same as text-based experiments. We use seen accuracy, unseen accuracy, harmonic mean of seen and unseen accuracy, and top-1 accuracy as the evaluation metrics.

\paragraph{Results:} In Table \ref{tb:awa2apysun}, we see that GRW outperforms all the existing methods on the seen-unseen harmonic mean for AwA2, aPY, and SUN datasets. In the case of the AwA2 dataset, it outperforms all the compared methods by a significant margin, i.e., 15.1\% in harmonic mean, and is also competent with existing methods in Top-1 accuracy while improving 4.8\%. GAZSL~\cite{Elhoseiny_2018_CVPR}+GRW has an average relative improvement  over GAZSL~\cite{Elhoseiny_2018_CVPR}+CIZSL~\cite{elhoseiny2019creativity} and GAZSL~\cite{Elhoseiny_2018_CVPR} of 24.92\% and 61.35\% in harmonic mean.
\begin{table*}[]
    \centering\footnotesize
    \begin{tabular}{@{}lccccccc@{}}
\toprule
 & \multicolumn{3}{c}{Top-1 Accuracy(\%)} &  & \multicolumn{3}{c}{Seen-Unseen H} \\ \cmidrule(lr){2-4} \cmidrule(l){6-8} 
 & AwA2 & aPY & SUN &  & AwA2 & aPY & SUN \\ \midrule
SJE~\cite{akata2015evaluation} & 61.9 & 35.2 & 53.7 &  & 14.4 & 6.9 & 19.8 \\
LATEM~\cite{xian2016latent} & 55.8 & 35.2 & 55.3 &  & 20.0 & 0.2 & 19.5 \\
ALE~\cite{akata2016label} & 62.5 & 39.7 & 58.1 &  & 23.9 & 8.7 & 26.3 \\
SYNC~\cite{changpinyo2016synthesized} & 46.6 & 23.9 & 56.3 &  & 18.0 & 13.3 & 13.4 \\
SAE~\cite{kodirov2017semantic} & 54.1 & 8.3 & 40.3 &  & 2.2 & 0.9 & 11.8 \\
DEM~\cite{zhang2016learning} & 67.1 & 35.0 & 61.9 &  & 25.1 & 19.4 & 25.6 \\
FeatGen~\cite{xian_feature_2018} & 54.3 & 42.6 & 60.8 &  & 17.6 & 21.4 & 24.9 \\
cycle-(U)WGAN~\cite{felix2018multi} & 56.2 & 44.6 & 60.3 &  & 19.2 & 23.6 & 24.4 \\ \midrule
LsrGAN (\textit{tr})~\cite{vyas2020leveraging} & $60.1$ & $34.6$ & 62.5 & & $48.7$ & $31.5$ & 44.8\\
 \quad + {\textbf{GRW}}& 63.7\textcolor{blue}{$^{+3.6}$} & 35.5\textcolor{blue}{$^{+0.9}$} & \textbf{64.2}\textcolor{blue}{$^{+1.7}$} & & \textbf{49.2}\textcolor{blue}{$^{+0.5}$} & \textbf{32.7}\textcolor{blue}{$^{+1.2}$} & \textbf{46.1}\textcolor{blue}{$^{+1.3}$} \\ \midrule
GAZSL~\cite{Elhoseiny_2018_CVPR} & 58.9 & 41.1 & 61.3 &  & 15.4 & 24.0 & 26.7 \\ 
\quad + {CIZSL}~\cite{elhoseiny2019creativity} & {67.8} & {42.1}& 63.7  &  & 24.6 & {25.7} & {27.8}  \\
\quad + {\textbf{GRW}} & \textbf{68.4}\textcolor{blue}{$^{+9.5}$} & \textbf{43.3}\textcolor{blue}{$^{+2.2}$} & 62.1\textcolor{blue}{$^{+0.8}$} &  & 39.0\textcolor{blue}{$^{+23.6}$} & 27.2\textcolor{blue}{$^{+3.2}$} & 27.9\textcolor{blue}{$^{+1.2}$} \\ 
\bottomrule
\end{tabular}  
\caption{Zero-Shot Recognition on class-level attributes of \textbf{AwA2}, \textbf{aPY} and \textbf{SUN} datasets, showing that GRW loss can improve the performance on attribute-based datasets. }\label{tb:awa2apysun}
\end{table*}

\section{Continual zero-shot learning experiments} \label{appen:czsl}
\subsection{Dataset and Continual Zero-Shot Learning Setup}
We display the seen and unseen class conversions in each task for each dataset in the Table \ref{tab:setting} to provide a better understanding of the specific implementation of CZSL on different datasets. Covered class means the number of unseen class converted to seen class per task. 
\begin{table}[]
\footnotesize
\minipage[t]{0.4\textwidth}\centering%
{%
\begin{tabular}{ccccc}
\toprule
 & AWA1 & AWA2 & CUB & SUN \\ \midrule
Total classes & 50 & 50 & 200 & 705 \\
Number of tasks & 5 & 5 & 20 & 15 \\
Initial seen classes & 10 & 10 & 10 & 47 \\
Covered class & 10 & 10 & 10 & 47 \\ \bottomrule
\end{tabular}\caption{Seen and Unseen classes in different dataset}
\label{tab:setting}
}
\endminipage\hfill
\minipage[t]{0.56\textwidth}\centering%
{%
\begin{tabular}{ccclcclcclcc}
\toprule
 & \multicolumn{2}{c}{AWA1} &  & \multicolumn{2}{c}{AWA2} &  & \multicolumn{2}{c}{CUB} &  & \multicolumn{2}{c}{SUN} \\ \cmidrule{2-3} \cmidrule{5-6} \cmidrule{8-9} \cmidrule{11-12} 
 & Inter. & Dic. &  & Inter. & Dic. &  & Inter. & Dic. &  & Inter. & Dic.\\ \midrule
$\lambda_\text{c}$ & 10 & 1 &  & 1 & 10 &  & 1 & 1 &  & 1 & 1 \\
$\lambda_\text{i}$ & 0.5 & 2 &  & 1 & 5 &  & 2 & 2 &  & 5 & 1 \\
$R$ & 3 & 3 &  & 3 & 3 &  & 5 & 5 &  & 5 & 5 \\ \bottomrule
\end{tabular}}
\caption{The hyperparameter for Table 1}
\label{tab:mainhyper}%
\endminipage\hfill
\end{table}


\subsection{More Ablations}

\begin{figure}[h!]
    \centering
    \includegraphics[width=0.8\linewidth]{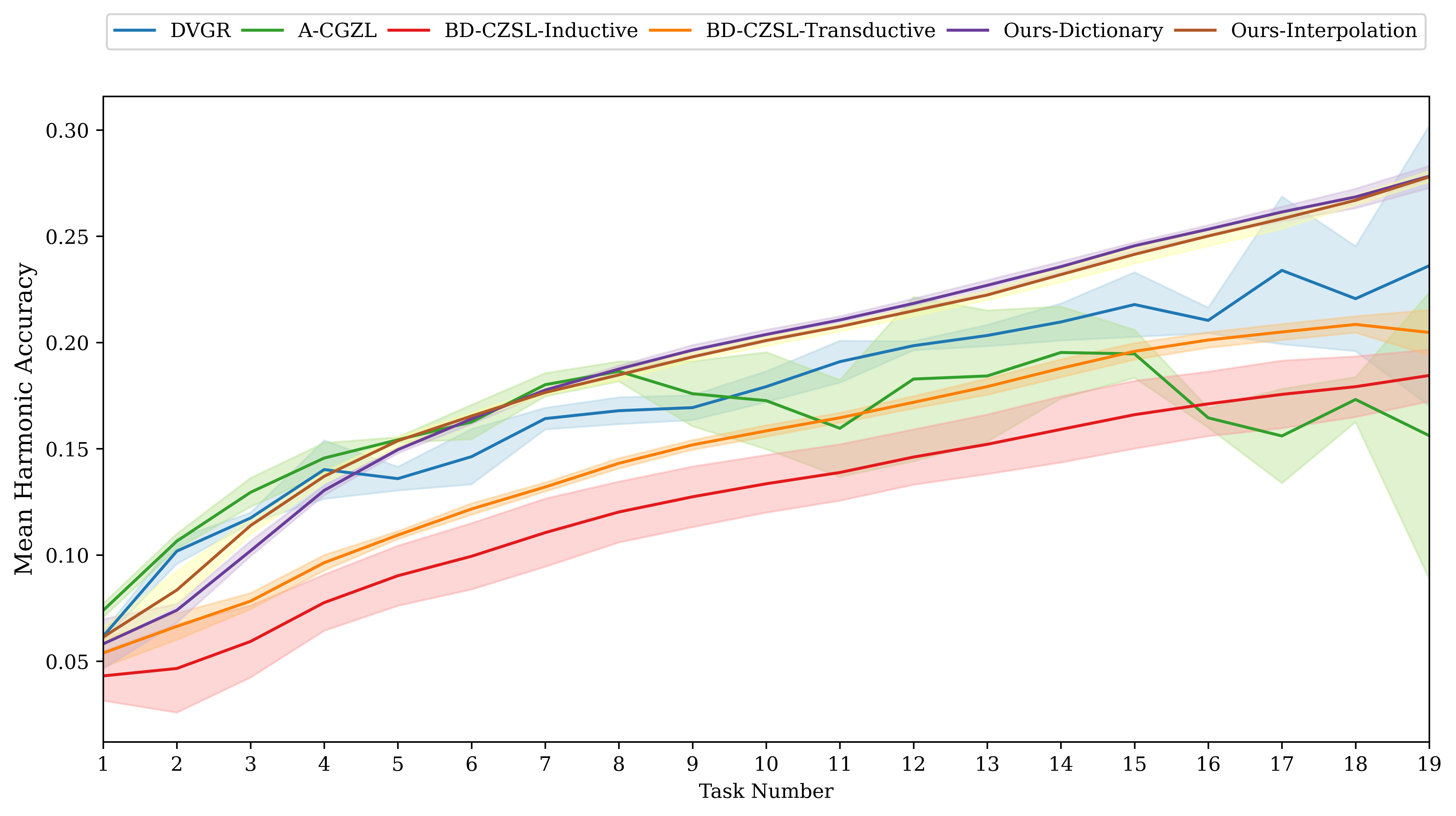}
    \caption{Mean harmonic accuracy at the end of each task with 5 different random seeds on CUB~\cite{WahCUB_200_2011}. Lines show the mH, and shadows show the standard deviation.}
    \label{fig:seed}
\end{figure}

\paragraph{Random seed:}
We experiment with multiple random seeds on the CUB dataset and show the averaged mH (line) and standard deviation (shadow) in Figure \ref{fig:seed}. The random seed mainly affects the generation part of GZSL learners. The generated data is used directly or indirectly to train the classifier of the unseen class. Figure \ref{fig:seed} shows that previous models are sensitive to random seeds, but our model is not. Previous models use the generated data as replay data or directly train the classifier, while ours avoids these. Our method uses a non-parametric classifier, a similarity-based classifier. During training, we pay more attention to improving the generalization ability of our embedder (discriminator) by encouraging the consistency between the generated visual space and the true visual space. Plus, we store the real data in the buffer. These all make our model more stable. Although we only reported the results of one seed (2222) in Table 4, the figure shows that the effect of different seeds on the results is not significant.

We also report mean and standard deviation of multiple runs of our methods in each dataset in Table \ref{tab:random_seed_inter}, \ref{tab:random_seed_dict}. It shows that experiments on all the datasets with both attribute generation  methods have relatively small variance. Although interpolation-based method has lower mean harmonic accuracy on fine-grained dataset CUB and SUN, it is shown to be more stable with less variance than dictionary-based method.  
\begin{table*}[]\centering\footnotesize
\minipage[t]{0.48\textwidth}\centering
{
\begin{tabular}{ccccccccc}
\toprule
 & \multicolumn{2}{c}{mSA} &  & \multicolumn{2}{c}{mUA} &  & \multicolumn{2}{c}{mHA} \\  \midrule
 & Mean & Std  &  & Mean & Std&  & Mean & Std  \\ \cmidrule{2-3} \cmidrule{5-6} \cmidrule{8-9} 
AWA1 & 65.87 & 1.19 &  & 33.77 & 1.00 &  & 42.69 & 0.57 \\
AWA2 & 70.52 & 0.46 &  & 34.52 & 0.90 &  & 44.45 & 0.79 \\
CUB & 42.11 & 0.88 &  & 22.10 & 0.67 &  & 27.80 & 0.53 \\
SUN & 36.29 & 0.18 &  & 21.07 & 0.33 &  & 26.44 & 0.20 \\ \bottomrule
\end{tabular}}
\caption{Our method in continual zero shot learning  with interpolated attributes. Mean and variance calculated on three runs with different random seeds.}
\label{tab:random_seed_inter}
\endminipage\hfill
\minipage[t]{0.48\textwidth}\centering%
{
\begin{tabular}{ccccccccc}
\toprule
 & \multicolumn{2}{c}{mSA} &  & \multicolumn{2}{c}{mUA} &  & \multicolumn{2}{c}{mHA} \\  \midrule
 & Mean & Std  &  & Mean & Std  &  & Mean & Std  \\ \cmidrule{2-3} \cmidrule{5-6} \cmidrule{8-9} 
AWA1 & 66.35 & 0.28 &  & 32.75 & 0.94 &  & 41.90 & 0.91 \\
AWA2 & 70.55 & 0.51 &  & 33.88 & 0.60 &  & 43.49 & 0.88 \\
CUB & 42.22 & 0.30 &  & 22.78 & 0.91 &  & 28.09 & 0.68 \\
SUN & 36.63 & 0.12 &  & 21.39 & 0.47 &  & 26.79 & 0.37 \\  \bottomrule
\end{tabular}}
\caption{Our method in inductive continual zero shot learning with learnable dictionary of attributes. Mean and variance calculated using three runs with different random seeds}
\label{tab:random_seed_dict}
\endminipage\hfill
\end{table*}

\subsection{Hyperparameters in GRW loss}\label{appen:exp_hyperparameters}

\textbf{Hyperparameter for Table 1:} We use the validation set to tune the hyperparameter random walk step $R$, coefficient of $L_\text{creativity}$ $\lambda_\text{c}$, and coefficient of $L_\text{inductive}$ $\lambda_\text{i}$. The hyperparameter used to report Table 1 is shown in Table \ref{tab:mainhyper}

\textbf{Walk length $R$ and decay rate $\gamma$:} We do an ablation study on the random walk length $R$ and decay rate $\gamma$ of the GRW loss in continual zero-shot learning experiments. Table \ref{tab:nrw} shows our method with different random walk lengths in AWA1 dataset and CUB dataset. 
In the dataset AWA1, moderate lengths give the highest mHA while in the CUB dataset higher random walk lengths provide the best mHA. It shows that the more challenging the dataset, the more random walk length is needed.
Unlike ZSL experiments, in CZSL experiments, knowledge is not only transferred to the unseen class space but also to the next task. Long walk length could give the model a more holistic view of the current task, but may harm the transformation to the next task. Therefore, tuning the number of random walk steps is required for new datasets. 

\begin{table*}[t]
\centering\footnotesize
\minipage[t]{0.49\textwidth}\centering%
\begin{tabular}{@{}lrrrrlrrr@{}}
\toprule
 & \multicolumn{1}{l}{} & \multicolumn{3}{c}{Ours Interpolation} &  & \multicolumn{3}{c}{Ours Dictionary} \\ \cmidrule(lr){3-5} \cmidrule(l){7-9} 
 & \multicolumn{1}{l}{} & \multicolumn{1}{l}{mSA} & \multicolumn{1}{l}{mUA} & \multicolumn{1}{l}{mHA} &  & \multicolumn{1}{l}{mSA} & \multicolumn{1}{l}{mUA} & \multicolumn{1}{l}{mHA} \\ \midrule
\multirow{3}{*}{$R$} & 1 & 41.7 & 21.2 & 27.1 &  & 43.6 & 22.7 & 28.4 \\
 & 3 & 42.3 & 22.1 & 27.7 &  & 42.1 & 20.6 & 26.6 \\
 & 5 & 42.2 & 22.7 & \textbf{28.4} &  & 42.4 & 23.6 & \textbf{28.8} \\ \bottomrule
\end{tabular}
\endminipage\hfill
\minipage[t]{0.49\textwidth}\centering%
\begin{tabular}{@{}ccccccccc@{}}
\toprule
 &  & \multicolumn{3}{c}{Ours Interpolation} &  & \multicolumn{3}{c}{Ours Dictionary} \\ \cmidrule(lr){3-5} \cmidrule(l){7-9} 
 &  & mSA & mUA & mHA &  & mSA & mUA & mHA \\ \midrule
\multirow{3}{*}{$R$} & 1 & 65.4 & 33.9 & 42.7 &  & 66.6 & 32.7 & 41.5 \\
 & 3 & 67.0 & 34.2 & \textbf{43.4} &  & 67.1 & 33.5 & \textbf{42.8} \\
 & 5 & 65.8 & 33.0 & 42.1 &  & 66.8 & 32.7 & 41.7 \\ \bottomrule
\end{tabular}%
\endminipage\hfill
\caption{Our method with different random walk length $R$ in AWA1 dataset (right) and CUB dataset (left)}\label{tab:nrw}
\end{table*}

Decay rate $\gamma$ works as a scale factor to the GRW loss to prevent a specific area in the probability matrix from being too close to one, resulting in exponential growth in the multiplication results when compared to other areas. Compared to the non-decay case when $\gamma = 1$ in Table \ref{tab:gamma}, the decayed case has noticeable improvements in unseen accuracy, resulting in better harmonic accuracy.
\begin{table*}
\centering\footnotesize
\minipage[t]{0.49\textwidth}\centering%
\begin{tabular}{ccccccccc}
\toprule
 &  & \multicolumn{3}{c}{Ours-interpolation} &  & \multicolumn{3}{c}{Ours-dictionary} \\ \cmidrule{3-5} \cmidrule{7-9} 
 &  & mSA & mUA &  mH &  & mSA & mUA &  mH \\ \midrule
$\gamma$ & 0.7 & 40.97 & 21.78 &  27.26 &  & 42.22 & 22.03 &  27.47 \\
 & 1 & 40.95 & 21.21 &  27.05 &  & 42.62 & 21.6 &  27.43 \\ \bottomrule
\end{tabular}%
\endminipage\hfill
\minipage[t]{0.49\textwidth}\centering%
\begin{tabular}{@{}clcrrlcrr@{}}
\toprule
\multicolumn{1}{l}{} &  & \multicolumn{3}{c}{Ours-interpolation} &  & \multicolumn{3}{c}{Ours-dictionary} \\ \cmidrule(lr){3-5} \cmidrule(l){7-9} 
 &  & mSA & \multicolumn{1}{c}{mUA} & \multicolumn{1}{c}{mH} &  & mSA & \multicolumn{1}{c}{mUA} & \multicolumn{1}{c}{mH} \\ \midrule
 & 0.7 & \multicolumn{1}{r}{66.8} & 33.42 &  42.87 &  & \multicolumn{1}{r}{66.93} & 32.41 &  41.51 \\
\multirow{-2}{*}{$\gamma$} & 1 & \multicolumn{1}{r}{66.07} & 32.31 &  41.69 &  & \multicolumn{1}{r}{66.34} & 32.87 &  41.76 \\ \bottomrule
\end{tabular}%
\endminipage\hfill
\caption{Our method with different decay rate $\gamma$ on CUB dataset (left) and AWA1 dataset (right)}\label{tab:gamma}
\end{table*}

\subsection{Ablations on Weight of Inductive Loss }
\paragraph{Inductive weight $\lambda_i$}
We also do an ablation study on the inductive coefficient $\lambda_i$ in Table  \ref{tab:induccoef}. This factor mainly affects the proportion of inductive loss in the overall loss. We found that our model is not sensitive to this hyperparameter. Whether on the larger dataset CUB or the smaller dataset AWA1, the difference of mH of different $\lambda_i$ on our model does not exceed 1\%. Therefore, our model does not need too much parameter tuning process.
\begin{table*}\centering\footnotesize
\minipage[t]{0.49\textwidth}\centering%
\begin{tabular}{@{}ccccclccc@{}}
\toprule
 &  & \multicolumn{3}{c}{Ours-interpolation} &  & \multicolumn{3}{c}{Ours-dictionary} \\ \cmidrule(lr){3-5} \cmidrule(l){7-9} 
 &  & mSA & mUA & mH &  & mSA & mUA & mH \\ \midrule
 & 0.01 & 41.81 & 20.93 &  27.01 &  & 42.8 & 23.07 &  28.51 \\
 & 0.1 & 42.32 & 21.27 &  27.11 &  & 42.73 & 21.98 &  27.85 \\
\multirow{-3}{*}{$\lambda_i$} & 1 & 40.97 & 21.78 &  27.26 &  & 42.22 & 22.03 &  27.47 \\ \bottomrule
\end{tabular}%
\endminipage\hfill
\minipage[t]{0.49\textwidth}\centering%
\begin{tabular}{@{}clccclccc@{}}
\toprule
\multicolumn{1}{l}{} &  & \multicolumn{3}{c}{Ours-interpolation} &  & \multicolumn{3}{c}{Ours-dictionary} \\ \cmidrule(lr){3-5} \cmidrule(l){7-9} 
 &  & mSA & mUA & mH &  & mSA & mUA & mH \\ \midrule
 & 0.1 & 66.81 & 32.82 &  42.15 &  & 66.32 & 32.11 &  41.15 \\
 & 1 & 66.8 & 33.42 &  42.87 &  & 66.93 & 32.41 &  41.51 \\
\multirow{-3}{*}{$\lambda_i$} & 10 & 66.38 & 33.77 &  42.92 &  & 66.47 & 31.81 &  40.89 \\ \bottomrule
\end{tabular}%
\endminipage\hfill
\caption{Our method with different inductive coefficients $\lambda_i$ on CUB dataset (left) and AWA1 dataset (right)}\label{tab:induccoef}
\end{table*}


\subsection{Continual zero-shot learning with other common settings}\label{appen:comparegr}
Although our main research problem is inductive setting, and we think real replay is needed, we still have an open attitude to other settings and migrate our model naively to their setting. We show experiment results in these settings in Table \ref{tab:settingcomparison} and compare them with other methods.

\begin{table*}
\centering
\caption{Comparison of our inductive loss in other common CZSL settings}
\label{tab:settingcomparison}
\resizebox{\textwidth}{!}{%
\begin{tabular}{@{}ccccccccccccccc@{}}
\toprule
 & replay method & zsl setting & \multicolumn{3}{c}{AWA1} & \multicolumn{3}{c}{AWA2} & \multicolumn{3}{c}{CUB} & \multicolumn{3}{c}{SUN} \\ \cmidrule(l){4-15} 
\multicolumn{1}{l}{} & \multicolumn{1}{l}{} & \multicolumn{1}{l}{} & mSA & mUA & mHA & mSA & mUA & mHA & mSA & mUA & mHA & mSA & mUA & mHA \\ \midrule
CN-ZSL & real & in & - & - &  - & 33.55 & 6.44 &  10.77 & 44.31 & 14.8 &  22.7 & 22.18 & 8.24 &  12.46 \\
Ours-interpolation & real & in & 62.9 & 32.77 &  \textbf{42.03} & 67.41 & 35.4 &  \textbf{45.06} & 40.17 & 21.78 &  27.26 & 36.29 & 21.05 &  26.51 \\
Ours-dictionary & real & in & 63.43 & 32 &  41.15 & 68.02 & 33.22 &  42.89 & 41.45 & 22.03 &  \textbf{27.47} & 36.54 & 21.31 &  \textbf{26.76} \\ \midrule
DVGR & generative & tr & 65.1 & 28.5 &  38 & 73.5 & 28.8 &  40.6 & 44.87 & 14.55 &  21.66 & 22.36 & 10.67 &  14.54 \\
A-CGZSL & generative & tr & 70.16 & 25.93 &  37.19 & 70.16 & 25.93 &  37.19 & 34.25 & 12.42 &  17.41 & 17.2 & 6.31 &  9.68 \\
BD-CGZSL & generative & tr & 67.55 & 36.04 &  \textbf{47.88} & 71.37 & 38.76 &  \textbf{51.6} & 31 & 23.97 &  \textbf{26.01} & 30.08 & 20.07 &  23.72 \\
Ours-interpolation & generative & tr & 62.43 & 33.03 &  42.01 & 66.84 & 34.01 &  43.77 & 32.53 & 16.66 &  21.65 & - & - &  - \\
Ours-dictionary & generative & tr & 62.34 & 31.5 &  40.18 & 68.07 & 34.45 &  44.17 & 30 & 16.18 &  20.55 & - & - &  - \\ \midrule
BD-CGZSL-in & generative & in & 62.12 & 31.51 &  40.46 & 67.68 & 32.88 &  42.33 & 37.76 & 9.089 &  14.43 & 34.93 & 14.86 &  20.8 \\
Ours-interpolation & generative & in & 61.43 & 34.04 &  \textbf{42.18} & 67.34 & 35.29 &  \textbf{44.95} & 29.78 & 16.86 &  \textbf{21.06} & 30.9 & 18.4 &  \textbf{22.99} \\
Ours-dictionary & generative & in & 62.26 & 30.88 &  39.68 & 67.44 & 33.68 &  43.24 & 28.34 & 16.94 &  20.57 & 30.13 & 18.56 &  22.85 \\ \bottomrule
\end{tabular}%
}
\end{table*}
We mentioned earlier that the generative replay method has unbalanced storage and buffer overload problems, but many models still use generative replay. When data privacy concerns are encountered, the generative replay method may be an alternative to the real replay method.  When using the generative replay, our model outperforms most existing methods. Our problem analysis cannot be applied in this setting, since we believe the replayed feature should have a balanced number in each class. 

Our primary focus is on the inductive setting, but we also provide results in the transductive setting and with generative replay. In the transductive setting, we use the ground truth unseen attributes to generate the visual features, and our loss works on these generations. Our method is comparable with other transductive methods, even without carefully designing how to use the semantic information. 

Through these knots, we believe that our model has the possibility of being migrated to other settings and is valuable for further explorations in other settings.

\end{document}